\documentclass[11pt]{article}
\usepackage{amssymb}
\usepackage{amsmath}
\usepackage{amsfonts}
 \usepackage{pifont}
\usepackage{amsmath,amsfonts,amssymb,theorem,graphicx,listings,txfonts}
\usepackage{graphics}
 \usepackage{caption2}
\usepackage{flafter}
 \usepackage{indentfirst}
\usepackage{epsfig}
  \usepackage{mathrsfs}
\usepackage{subfigure}
 \usepackage{cite}
\newtheorem{theorem}{Theorem}[section]

\newtheorem{definition}[theorem]{Definition}

\theoremstyle{example}
\newtheorem{example}[theorem]{Example}
\theoremstyle{programme}

\theoremstyle{property}

\theoremstyle{problem}

\renewcommand{\arraystretch}{1}

\title{Decision-theoretic rough sets-based three-way approximations of interval-valued fuzzy sets}

\author
{Guangming Lang$^{a}$ 
\thanks{Corresponding author.\quad Tel./fax: +86 731 8822855,
\newline\mbox{}\hspace{0.55cm}
E-mail address: langguangming1984@126.com(G.M.Lang). }\hspace{1cm}
\\
\small {$^{a}$ School of Mathematics and Computer Science, Changsha University of Science and Technology}\\
\small {Changsha, Hunan 410082, P.R. China}\\
\small {$^{b}$ College of Mathematics and Econometrics, Hunan University}\\
\small {Changsha, Hunan 410082, P.R. China}\\
}
\date{}

\begin{document}
\maketitle \baselineskip=17pt
\begin{center}
\begin{quote}
{{\bf Abstract.}
In practical situations, interval-valued fuzzy sets
are frequently encountered. In this paper, firstly, we present
shadowed sets for interpreting and understanding interval fuzzy
sets. We also provide an analytic solution to
computing the pair of thresholds by searching for a balance
of uncertainty in the framework of shadowed sets.
Secondly, we construct errors-based three-way approximations of
interval-valued fuzzy sets. We also
provide an alternative decision-theoretic formulation for
calculating the pair of thresholds by transforming interval-valued loss functions into single-valued loss functions,
in which the required thresholds are computed by minimizing decision
costs. Thirdly, we compute errors-based three-way approximations of
interval-valued fuzzy sets by using interval-valued loss functions. Finally, we employ several examples to illustrate that how to
take an action for an object with interval-valued membership grade by using interval-valued loss
functions.

{\bf Keywords:} Decision-theoretic rough sets; Interval-valued fuzzy sets; Interval-valued loss function; Shadowed set
\\}
\end{quote}
\end{center}
\renewcommand{\thesection}{\arabic{section}}

\section{Introduction}

Interval-valued fuzzy sets\cite{Turksen1}, as an extension of fuzzy sets\cite{Zadeh1}, is
a powerful mathematical tool to describe uncertainty information, in
which the concept of the membership function using the subintervals
of the interval $[0,1]$ as the set of membership grades is a
fundamental notion. It has been intensively investigated, not only its
theoretical aspects, but also its numerous applications, and the approximations of interval-valued fuzzy sets by using several levels of membership grades have became an important research direction.

Recently,
researchers\cite{Banerjee1,Chakrabarty1,Deng1,Dubois1,Grzegorzewski1,
Klir1,Pedrycz1,Liang1,Liang2,Liu1,Moore1,Nasibov1,Zadeh2} have
investigated fuzzy sets from different aspects. For example,
Pedrycz\cite{Pedrycz1} proposed shadowed sets for interpreting fuzzy
sets by using several levels of membership grades, in which, if the
membership grade of an element is close to $1$, it would be
considered to be the same as $1$ and elevated to $1$; If the
membership grade is close to $0$, it would be considered to be the
same as $0$ and is reduced to $0$; If the membership grade is
neither close to $0$ nor close to $1$, it would be put into a
shadowed region, in which the elevation and reduction operations use
thresholds that provide semantically meaningful and acceptance
levels of degree of closeness of membership values to $1$ and $0$,
respectively. Sequently, a lot of
investigations\cite{Pedrycz2,Pedrycz3,Pedrycz4,Pedrycz5,Pedrycz6,
Cattaneo1,Cattaneo2,Cattaneo3,Deng2,
Grzegorzewski2,Mitra1,Wang1,Zhou1} have been done on shadowed sets.
For instance, Deng et
al.\cite{Deng1} computed a pair of thresholds, whose interpretation
and determination is a fundamental issue for expressing fuzzy sets,
for three-way approximations of fuzzy sets by using loss functions,
and classify a set of objects into three regions by using the pair
of thresholds. In practical situations, interval-valued fuzzy sets whose
membership functions are using the subintervals of the interval
$[0,1]$ are of interest because such type of sets are frequently
encountered. So far we have not seen the similar investigation on
interval-valued fuzzy sets. Therefore, it is of interest to investigate
that how to express interval-valued fuzzy sets as fuzzy sets.

To computing and interpreting a pair of thresholds, a lot of
investigations\cite{Li1,Liang1,Liang2,Yao1,Yao2,Yao3,Yao4,Yao5,
Liu1,Liu2, Liu3} have been done on three-way decision-theory by
using loss functions in literatures. For example, Li et
al.\cite{Li1} evaluated the cost and benefit of assigning an
instance to a specific subcategory and defined a general loss
function for supervised leaning. Liang et al.\cite{Liang1,Liang2}
presented triangular fuzzy decision-theoretic rough sets and
systematic studies on three-way decisions with interval-valued
decision-theoretic rough sets. Liu et al.\cite{Liu2} proposed
stochastic decision-theoretic rough sets, interval-valued
decision-theoretic rough sets, fuzzy decision-theoretic rough sets
and dynamic decision-theoretic rough sets. In practical situations,
interval-valued loss functions as interval-valued
numbers\cite{Qian1,Sengupta1,Liu1} are of interest because such
functions are frequently encountered. Although interval-valued loss
functions are complex in practice, we have not seen enough
investigations on interval-valued fuzzy sets by using interval-valued loss
functions so far. Therefore, it is urgent to further study interval-valued
loss functions for making a decision by using three-way
decision-theory.

The purpose of this paper is to further investigate
interval-valued fuzzy sets. Section 2 introduces the
basic principles of decision-theoretic rough sets, shadowed sets and decision-theoretic three-way approximations of fuzzy sets.
Section 3
presents shadowed sets of interval-valued fuzzy sets.
Section 4 is devoting to
errors-based interpretation of shadowed sets of interval-valued fuzzy sets. Section 5 presents
decision-theoretic rough sets-based three-way approximations
of interval-valued fuzzy sets by using transforming interval-valued loss functions into single loss functions. Section 6 investigates decision-theoretic rough sets-based three-way approximations
of interval-valued fuzzy sets by using interval-valued loss functions from another view. The
conclusion comes in Section 7.

\section{Preliminaries}

In this section, we review some concepts of fuzzy sets, interval-valued fuzzy sets,
shadowed sets and decision-theoretic three-way approximations of
fuzzy sets.

\subsection{Shadowed sets of fuzzy sets}

In \cite{Zadeh1}, Zadeh presented the concept of fuzzy sets for interpreting uncertainty problems.

\begin{definition}\cite{Zadeh1}
Let $\mu_{A}$ be a mapping from $U$ to $[0,1]$ such as $\mu_{A}:
U\longrightarrow [0,1]:$ $ x\longrightarrow \mu_{A}(x),$ where $x\in
U$, $\mu_{A}$ is the membership function. Then $A$ is referred to as
a fuzzy set.
\end{definition}

In \cite{Pedrycz1}, Pedrycz presented the concept of shadowed sets for expressing fuzzy sets.

\begin{definition}\cite{Pedrycz1}
Let $A$ be a fuzzy set, the shadowed set $S_{\mu_{A}}$ of
$A$ is defined as \makeatother
$$S_{\mu_{A}}(x)=\left\{
\begin{array}{ccc}
1,&{\rm }& \mu(x)\geq \alpha;\\
0,&{\rm }& \mu(x)\leq \beta;\\
$[$0,1$]$,&{\rm }& \beta<\mu(x)< \alpha.
\end{array}
\right. $$
\end{definition}

In Pedrycz's model, an optimal pair of thresholds is computed by minimizing the absolute difference as
\begin{eqnarray*}
V_{(\alpha,\beta)}(\mu_{A})&=&|\text{Elevated
Area}_{(\alpha,\beta)}(\mu_{A}) +\text{Reduced
Area}_{(\alpha,\beta)}(\mu_{A})-\text{Shadowed
Area}_{(\alpha,\beta)}(\mu_{A})|\\
&=&|\sum_{\mu_{A}(x)\geq\alpha}(1-\mu_{A}(x))+
\sum_{\mu_{A}(x)\leq\beta}(\mu_{A}(x))-Card(\{x\in
U|\beta<\mu_{A}(x)<\alpha\})|,
\end{eqnarray*}
where card$(\cdot)$ denotes the cardinality of a set $\cdot$, and an
optimal pair of thresholds $\alpha$ and $\beta$ can be derived by
minimizing the objective function $V_{(\alpha,\beta)}(\mu_{A})$. Similarly, it is also difficult to compute the pair of
thresholds $\alpha$ and $\beta$ since minimizing
$V_{(\alpha,\beta)}(\mu_{A})$ involves two parameters $\alpha$ and
$\beta$. For convenience, by using $\alpha+\beta=1$, the objective
function is simplified into
\begin{eqnarray*}
V_{(\alpha,1-\alpha)}(\mu_{A})&=&|\text{Elevated
Area}_{(\alpha,1-\alpha)}(\mu_{A}) +\text{Reduced
Area}_{(\alpha,1-\alpha)}(\mu_{A})-\text{Shadowed
Area}_{(\alpha,1-\alpha)}(\mu_{A})|\\
&=&|\sum_{\mu_{A}(x)\geq\alpha}(1-\mu_{A}(x))+
\sum_{\mu_{A}(x)\leq1-\alpha}(\mu_{A}(x))-Card(\{x\in
U|1-\alpha(<\mu_{A}(x)<\alpha\})|.
\end{eqnarray*}

\subsection{Decision-theoretic three-way approximations of fuzzy sets}

In terms of the errors, Deng et al.\cite{Deng1} expressed the
objective function to further investigate shadowed sets of fuzzy sets as
\begin{eqnarray*}
V_{(\alpha,\beta)}(\mu_{A})&=&|E_{e}(\mu_{A})
+E_{r}(\mu_{A})-E_{s}(\mu_{A})|\\
&=&|\sum_{\mu_{A}(x)\geq\alpha}(1-\mu_{A}(x))+
\sum_{\mu_{A}(x)\leq\beta}(\mu_{A}(x))-\sum_{\beta<\mu_{A}(x)<
\alpha}(1-\mu_{A}(x))+\sum_{\beta<\mu_{A}(x)< \alpha}(\mu_{A}(x))|.
\end{eqnarray*}


The objective
function is constructed on elevated area, reduced
area and shadowed
area, and it
is necessary to investigate that which numeric value is meaningful
to the membership grade of elements in the shadowed area.

By replacing
the unit interval $[0,1]$ with $0.5$, Deng et al. provided
\begin{eqnarray*}
\makeatother T_{\mu_{A}}(x)=\left\{
\begin{array}{ccc}
1,&{\rm }& \mu(x)\geq \alpha;\\
0,&{\rm }& \mu(x)\leq \beta;\\
0.5,&{\rm }& \beta<\mu(x)< \alpha.
\end{array}
\right.
\end{eqnarray*}
Subsequently,
by analyzing $T_{\mu_{A}}(x)$,
we have
\begin{eqnarray*}
E_{(\alpha,\beta)}(\mu_{A})&=&E_{e}(\mu_{A})
+E_{r}(\mu_{A})+E_{s_{0.5}}(\mu_{A})\\
&=&\sum_{\mu_{A}(x)\geq\alpha}(1-\mu_{A}(x))+
\sum_{\mu_{A}(x)\leq\beta}(\mu_{A}(x))-\sum_{0.5<\mu_{A}(x)<
\alpha(t)}(\mu_{A}(x)-0.5)+\sum_{\beta<\mu_{A}(x)<
0.5}(0.5-\mu_{A}(x)).
\end{eqnarray*}

Correspondingly,
the total error as the summation of errors of all objects are expressed
as
\begin{eqnarray*}
E_{(\alpha,\beta)}(\mu_{A})&=&\sum_{x\in
U}E_{(\alpha,\beta)}(\mu_{A}(x)),
\end{eqnarray*}
where
\begin{eqnarray*}
\makeatother E_{(\alpha,\beta)}(\mu_{A}(x))=\left\{
\begin{array}{ccc}
1-\mu(x),&{\rm }& \mu(x)\geq \alpha;\\
0.5-\mu(x),&{\rm }& \beta<\mu(x)\leq0.5 ;\\
\mu(x)-0.5,&{\rm }& 0.5<\mu(x)< \alpha;\\
\mu(x)-0,&{\rm }& \mu(x)\leq \beta.
\end{array}
\right.
\end{eqnarray*}

The total error is minimized by minimizing the error of each
individual object, and we search for a pair of thresholds $\alpha$
and $\beta$ such that $E_{(\alpha,\beta)}(\mu_{A}(x))$ is minimized
for each object. We consider the following actions and associated
errors for minimizing the error of each object:
\begin{eqnarray*}
(1):\text{elevate to } 1: 1-\mu_{A}(x); (2):\text{reduce to } 0:
\mu_{A}(x)-0; (3):\text{reduce or elevate to } 0.5: |\mu_{A}(x)-0.5|.
\end{eqnarray*}

That is, the absolute differences between $\mu_{A}(x)$ and three
values 1, 0.5 and 0, respectively, are the associated errors. A
minimized difference is obtained if $\mu_{A}(x)$ is changed into a
value that is closest to $\mu_{A}(x)$.

\begin{table}[htbp]\renewcommand{\arraystretch}{1.5}
\caption{Loss function.}
 \tabcolsep0.14in
\begin{tabular}{c c c c c}
\hline  Action & \text{Fuzzy set membership grade} &\text{Three-way membership grade}& Error &Loss\\
\hline
$a_{e}$ & $\mu_{A}(x)\geq \alpha$& $1$  &$1- \mu_{A}(x)$&$\lambda_{e}$\\
$a_{r}$ & $\mu_{A}(x)\leq \beta$& $0$ & $\mu_{A}(x)$&$\lambda_{r}$\\
$a_{s_{\downarrow}}$ & $0.5\leq\mu_{A}(x)<\alpha$& $0.5$ &$\mu_{A}(x)-0.5$ &$\lambda_{s_{\downarrow}}$ \\
$a_{s_{\uparrow}}$ & $\beta<\mu_{A}(x)<0.5$& $0.5$ &$0.5-\mu_{A}(x)$ & $\lambda_{s_{\uparrow}}$\\
\hline
\end{tabular}
\end{table}

By considering various costs of the actions of
elevation and reduction, Deng et al. presented an analytic solution
of computing the pair of thresholds $\alpha$ and $\beta$ by using
loss functions.
In Table $2$, the set of actions
$\{a_{e},a_{r}, a_{s_{\downarrow}},a_{s_{\uparrow}}\}$
describes four possible actions on changing the membership grade.
The fuzzy membership grade $\mu_{A}(x)$
represents the state of object in the second column, and the errors of different actions
are given in the fourth column, and the losses of different actions
are given in the fifth column.

Suppose $
\lambda_{e}>0,\lambda_{r}>0,\lambda_{s_{\downarrow}}>0,\lambda_{s_{\uparrow}}>0,
 \lambda_{s_{\downarrow}}\leq\lambda_{r}$ and $
\lambda_{s_{\uparrow}}\leq\lambda_{e}$, we immediately have three
rules as (E) If $\mu_{A}(x)\geq \alpha$, then $T_{\mu_{A}}(x)=1$;
(R) If $\mu_{A}(x)\leq\beta$, then $T_{\mu_{A}}(x)=0$; (S) If
$\beta<\mu_{A}(x)<\alpha$, then $T_{\mu_{A}}(x)=0.5$, where
\begin{eqnarray*}
\alpha=\frac{2\lambda_{e}+\lambda_{s_{\downarrow}}}{2(\lambda_{e}+\lambda_{s_{\downarrow}})}
\text{ and
}\beta=\frac{\lambda_{s_{\uparrow}}}{2(\lambda_{r}+\lambda_{s_{\uparrow}})}.
\end{eqnarray*}

\section{Shadowed sets of interval-valued fuzzy sets and its errors-based interpretations }

In this section, we present the concept of shadowed sets of interval-valued
fuzzy sets and its errors-based interpretations for illustrating
interval-valued fuzzy sets.

\subsection{Shadowed sets of interval-valued fuzzy sets}

In this subsection, we present the concept of shadowed sets of
interval-valued fuzzy sets.

\begin{definition}
Let $D_{[0,1]}$ be the set of closed subintervals of the interval
$[0,1]$. An interval-valued fuzzy set $A$ in $U$ is given by
$A=\{(x,\widetilde{\mu}_{A}(x))|x\in U\}$, where $ \widetilde{\mu}_{A}: X\longrightarrow
D_{[0,1]}: x\longrightarrow
\widetilde{\mu}_{A}(x)=[\mu^{-}_{A}(x),\mu^{+}_{A}(x)]. $
\end{definition}

\begin{definition}
Let $A$ be an interval-valued fuzzy set,
$\widetilde{\mu}_{A}(x)=[\mu^{-}_{A}(x),\mu^{+}_{A}(x)]$ be a membership grade
of $x\in U$, and $\theta\in [0,1]$. Then the transformed formula of
$\widetilde{\mu}_{A}(x)$ is
$m_{\theta}(\widetilde{\mu}_{A}(x))=(1-\theta)\cdot\mu^{-}_{A}(x)+\theta\cdot\mu^{+}_{A}(x);$
Furthermore, $A_{\theta}=\{(x,m_{\theta}(\widetilde{\mu}_{A}(x)))|x\in U\}.$
\end{definition}

\begin{example}
$(1)$ Let $\widetilde{\mu}_{A}(x)=[0.1,0.2]$ and $\widetilde{\mu}_{B}(x)=[0.15,0.25]$ for $x\in U$, and $\theta=0.5$. Then \begin{eqnarray*}
m_{\theta}(\widetilde{\mu}_{A}(x))&=&(1-0.5)\times 0.1+0.5\times 0.2=0.15;\\
m_{\theta}(\widetilde{\mu}_{B}(x))&=&(1-0.5)\times 0.15+0.5\times 0.25=0.2.
\end{eqnarray*}

$(2)$ Let $A=\frac{x_{1}}{\widetilde{\mu}_{A}(x_{1})}+\frac{x_{2}}{\widetilde{\mu}_{A}(x_{2})}+\frac{x_{3}}{\widetilde{\mu}_{A}(x_{3})}+\frac{x_{4}}{\widetilde{\mu}_{A}(x_{4})}$ be an interval-valued fuzzy set, where
$\widetilde{\mu}_{A}(x_{1})=[0.1,0.2],\widetilde{\mu}_{A}(x_{2})=[0.6,0.8],\widetilde{\mu}_{A}(x_{3})=[0.3,0.5]$ and $\widetilde{\mu}_{A}(x_{4})=[0.8,0.1].$ If we take $\theta=0.5$, then
$A_{\theta}=\frac{x_{1}}{0.15}+\frac{x_{2}}{0.7}+\frac{x_{3}}{0.4}+\frac{x_{4}}{0.45}.$
\end{example}

For simplicity, we denote $\widetilde{\mu}_{A}$ as $\mu_{A}$ in the following discussion.

\begin{definition}
Let $A$ be an interval-valued fuzzy set, then the shadowed set
$S_{\mu_{A}}$ of $A$ is defined as
 \makeatother
$$S_{\mu_{A}}(x)=\left\{
\begin{array}{ccc}
1,&{\rm }& m_{\theta}(\mu_{A}(x))\geq \alpha;\\
0,&{\rm }& m_{\theta}(\mu_{A}(x))\leq \beta;\\
$[$0,1$]$,&{\rm }& \beta<m_{\theta}(\mu_{A}(x))< \alpha.
\end{array}
\right. $$
\end{definition}

For an object $x$, we elevate the membership grade from $\mu_{A}(x)$
to $1$ if $m_{\theta}(\mu_{A}(x))\geq \alpha$; We reduce the
membership grade from $\mu_{A}(x)$ to 0 if
$m_{\theta}(\mu_{A}(x))\leq \beta$; We change the membership grade
from $\mu_{A}(x)$ to $[0,1]$ if
$\beta<m_{\theta}(\mu_{A}(x))<\alpha$.

The pair of thresholds $\alpha$ and $\beta$ are
important for computing three-way approximations of interval-valued fuzzy
sets. In what follows, we introduce a systematic way to compute the
pair of thresholds $\alpha$ and $\beta$ by minimizing an objective
function as
\begin{eqnarray*}
V_{(\alpha,\beta)}(A)&=&|\text{Elevated
Area}_{(\alpha,\beta)}(A) +\text{Reduced
Area}_{(\alpha,\beta)}(A)-\text{Shadowed
Area}_{(\alpha,\beta)}(\mu_{A})|\\
&=&|\sum_{m_{\theta}(\mu_{A}(x))\geq\alpha}(1-m_{\theta}(\mu_{A}(x)))+
\sum_{m_{\theta}(\mu_{A}(x))\leq\beta}(m_{\theta}(\mu_{A}(x)))-Card(\{x\in
U|\beta<m_{\theta}(\mu_{A}(x))<\alpha\})|,
\end{eqnarray*}
where card$(\cdot)$ denotes the cardinality of a set $\cdot$, and an
optimal pair of thresholds $\alpha$ and $\beta$ can be derived
by minimizing the objective function
$V_{(\alpha,\beta)}(A)$. Similarly,
minimizing $V_{(\alpha,\beta)}(A)$ involves two
parameters $\alpha$ and $\beta$. For convenience, by assuming
that $\alpha+\beta=1$, the objective function is simplified
into
\begin{eqnarray*}
&&V_{(\alpha,1-\alpha)}(A)\\&&=|\text{Elevated
Area}_{(\alpha,1-\alpha)}(m_{\theta}(\mu_{A}(x))) +\text{Reduced
Area}_{(\alpha,1-\alpha)}(m_{\theta}(\mu_{A}(x)))-\text{Shadowed
Area}_{(\alpha,1-\alpha)}(m_{\theta}(\mu_{A}(x)))|\\
&&=|\sum_{m_{\theta}(\mu_{A}(x))(x)\geq\alpha}(1-m_{\theta}(\mu_{A}(x)))+
\sum_{m_{\theta}(\mu_{A}(x))\leq1-\alpha}(m_{\theta}(\mu_{A}(x)))-Card(\{x\in
U|1-\alpha<m_{\theta}(\mu_{A}(x))<\alpha\})|.
\end{eqnarray*}

There exist two interpretations of shadowed sets of interval-valued fuzzy sets. In a wide
sense, a shadowed set is a three-valued fuzzy set,
which is used to approximate an interval-valued fuzzy set. In a narrow sense,
we interpret the notion of a shadowed set according to its
exact formulation, namely, the choice of the set of membership
grades $\{0,[0,1],1\}$  and the objective function.
Therefore,
shadowed sets of interval-valued fuzzy sets are examples of three-way approximations of
interval-valued fuzzy sets.

\subsection{Errors-based interpretation of shadowed sets for interval-valued fuzzy sets}

In this section, we present a detailed analysis of a objective
function for shadowed sets of interval-valued fuzzy sets in terms of errors of
approximations. We also provide a new objective function by the
total error of approximations for determining the thresholds
$\alpha$ and $\beta$.

To further study shadowed sets of interval-valued fuzzy sets, we express
the objective function in terms of the errors introduced by a
shadowed set approximation. For an object $x$ with membership grade
$m_{\theta}(\mu_{A}(x))$, the elevation operation changes the
membership grade from $m_{\theta}(\mu_{A}(x))$ to $1$, the reduction
operation changes the membership grade from $m_{\theta}(\mu_{A}(x))$
to $0$, and the errors induced by elevation and reduction are shown
as
\begin{eqnarray*}
E_{e}(\mu_{A}(x))=1-m_{\theta}(\mu_{A}(x)),
E_{r}(\mu_{A}(x))=m_{\theta}(\mu_{A}(x)).
\end{eqnarray*}

The errors $E_{e}(\mu_{A})$ and $E_{r}(\mu_{A})$ induced by the
elevation and reduction operations for an interval-valued fuzzy set $A$ of
the universe $U$, respectively, are shown as
\begin{eqnarray*}
E_{e}(\mu_{A})=\sum_{m_{\theta}(\mu_{A}(x))\geq \alpha}(1-m_{\theta}(\mu_{A}(x))),
E_{r}(\mu_{A}(x))=\sum_{m_{\theta}(\mu_{A}(x))\leq \beta}m_{\theta}(\mu_{A}(x)).
\end{eqnarray*}

The error for the shadowed area is not clear because of the unit
interval $[0, 1]$ as the membership grade when
$\beta<m_{\theta}(\mu_{A}(x))< \alpha$. By computing the difference
between $m_{\theta}(\mu_{A}(x))$ and the maximum $1$ and the minimum
value $0$ and summarizing them up, we have
\begin{eqnarray*}
E_{s}(\mu_{A})=\sum_{\beta<m_{\theta}(\mu_{A}(x))<
\alpha}(1-m_{\theta}(\mu_{A}(x)))+\sum_{\beta<m_{\theta}(\mu_{A}(x))<
\alpha}m_{\theta}(\mu_{A}(x)).
\end{eqnarray*}

Subsequently, we express the objective function in terms of
errors by using the error-based interpretation of the three areas
as
\begin{eqnarray*}
&&V_{(\alpha,\beta)}(\mu_{A})\\&&=|E_{e}(\mu_{A})
+E_{r}(\mu_{A})-E_{s}(\mu_{A})|\\
&&=|\sum_{m_{\theta}(\mu_{A}(x))\geq\alpha}(1-m_{\theta}(\mu_{A}(x)))+
\sum_{m_{\theta}(\mu_{A}(x))\leq\beta}m_{\theta}(\mu_{A}(x)-\sum_{\beta<m_{\theta}(\mu_{A}(x))<
\alpha}(1-m_{\theta}(\mu_{A}(x)))+\sum_{\beta<m_{\theta}(\mu_{A}(x))<
\alpha}(m_{\theta}\mu_{A}(x))|.
\end{eqnarray*}

The
objective function is a kind of trade-off of errors produced by
three regions. But the rationale for such a trade-off is not
entirely clear. On one hand, $E_{s}(m_{\theta}(\mu_{A}(x)))$
consists the errors of elevation and reduction operations,  and it
is impossible to elevate $m_{\theta}(\mu_{A}(x))$ to $1$ and reduce
$m_{\theta}(\mu_{A}(x))$ to $0$ if $\beta<m_{\theta}(\mu_{A}(x))<
\alpha$ simultaneously. On the other hand, we are not able to
allocate any numeric membership grade for the elements in the
shadowed area. In other words, any numeric value of the unit
interval $[0,1]$ could be permitted to reflect the uncertainty.
Therefore, it is necessary to investigate that which numeric value
is meaningful to the membership grade of elements in the shadowed
area.

Below, we present a three-way approximation of an interval-valued fuzzy
set by replacing the unit interval $[0,1]$ with $0.5$ as,
\begin{eqnarray*}
\makeatother T_{\mu_{A}}(x)=\left\{
\begin{array}{ccc}
1,&{\rm }& m_{\theta}(\mu_{A}(x))\geq \alpha;\\
0,&{\rm }& m_{\theta}(\mu_{A}(x))\leq \beta;\\
0.5,&{\rm }& \beta<m_{\theta}(\mu_{A}(x))< \alpha.
\end{array}
\right.
\end{eqnarray*}

By analyzing $T_{\mu_{A}}(x)$, we see that the correspondences
between areas of elevation and reduction and errors of elevation and
reduction remain to be the same. But we need to revise the errors of
the shadowed region as
\begin{eqnarray*}
E_{s_{0.5}}(\mu_{A})=\sum_{0.5<m_{\theta}(\mu_{A}(x))<
\alpha}(1-m_{\theta}(\mu_{A}(x)))+\sum_{\beta<m_{\theta}(\mu_{A}(x))<
0.5}(m_{\theta}(\mu_{A}(x))).
\end{eqnarray*}

By using $E_{e}(\mu_{A}),E_{r}(\mu_{A})$ and $E_{s_{0.5}}(\mu_{A})$,
we have
\begin{eqnarray*}
E_{(\alpha,\beta)}(\mu_{A})&=&E_{e}(\mu_{A})
+E_{r}(\mu_{A})+E_{s_{0.5}}(\mu_{A})\\
&=&\sum_{m_{\theta}(\mu_{A}(x))\geq\alpha}(1-m_{\theta}(\mu_{A}(x))+
\sum_{m_{\theta}(\mu_{A}(x))\leq\beta}(m_{\theta}(\mu_{A}(x)))-\sum_{0.5<m_{\theta}(\mu_{A}(x))<
\alpha(t)}(m_{\theta}(\mu_{A}(x))-0.5)\\&&+\sum_{\beta<m_{\theta}(\mu_{A}(x))<
0.5}(0.5-m_{\theta}(\mu_{A}(x))).
\end{eqnarray*}

The total errors of the three areas are minimized instead of
searching for a trade-off between different areas. Correspondingly,
we express the total error as the summation of errors of all objects
as
\begin{eqnarray*}
E_{(\alpha,\beta)}(\mu_{A})&=&\sum_{x\in
U}E_{(\alpha,\beta)}(m_{\theta}(\mu_{A}(x))),
\end{eqnarray*}
where
\begin{eqnarray*}
\makeatother E_{(\alpha,\beta)}(\mu_{A}(x))=\left\{
\begin{array}{ccc}
1-m_{\theta}(\mu_{A}(x)),&{\rm }& m_{\theta}(\mu_{A}(x))\geq \alpha;\\
0.5-m_{\theta}(\mu_{A}(x)),&{\rm }& \beta<m_{\theta}(\mu_{A}(x))\leq0.5 ;\\
m_{\theta}(\mu_{A}(x))-0.5,&{\rm }& 0.5<m_{\theta}(\mu_{A}(x))< \alpha;\\
m_{\theta}(\mu_{A}(x)),&{\rm }& m_{\theta}(\mu_{A}(x))\leq \beta.
\end{array}
\right.
\end{eqnarray*}

The total error will be minimized by minimizing the error of each
individual object, and we can search for a pair of thresholds
$\alpha$ and $\beta$ such that
$E_{(\alpha,\beta)}(\mu_{A}(x))$ is minimized for each
object. We consider the following actions and associated errors for
minimizing the error of each object:
\begin{eqnarray*}
(1):\text{elevate to } 1: 1-m_{\theta}(\mu_{A}(x)); (2):\text{reduce to } 0:
m_{\theta}(\mu_{A}(x))-0; (3):\text{reduce or elevate to } 0.5:
|m_{\theta}(\mu_{A}(x))-0.5|.
\end{eqnarray*}

That is, the absolute differences between $m_{\theta}(\mu_{A}(x))$ and three
values $1$, $0.5$ and $0$, respectively, are the associated errors.
A minimized difference is obtained if $m_{\theta}(\mu_{A}(x))$ is changed into
a value that is closest to $m_{\theta}(\mu_{A}(x))$.

\section{Decision-theoretic rough sets-based three-way approximations
of interval-valued fuzzy sets}

In this section, we introduce a framework for decision-theoretic
rough sets-based three-way approximations of interval-valued fuzzy sets.

\subsection{Cost-sensitive three-way approximations of interval-valued fuzzy sets}

In Section 3, we investigate three-way approximation of interval-valued
fuzzy sets by using three membership grades of $0$, $0.5$ and $1$.
We take one of the following three actions for an object with a
membership grade: elevate the membership grade to 1, reduce
the membership grade to 0, and change the membership grade to 0.5.
More specially, there are two situations for the third case: reduce
the membership grade to 0.5 if $m_{\theta}(\mu_{A}(x))\geq 0.5$ and elevate
the membership grade to 0.5 if $m_{\theta}(\mu_{A}(x))< 0.5$. Each action will
incur error and the costs of different actions are not necessarily
the same.

\begin{table}[htbp]\renewcommand{\arraystretch}{1.5}
\caption{Loss function.}
 \tabcolsep0.155in
\begin{tabular}{c c c c c}
\hline  Action & \text{Membership grade} &\text{Three-way membership grade}& Error &Loss\\
\hline
$a_{e}$ & $m_{\theta}(\mu_{A}(x))\geq \alpha$& $1$  &$1- m_{\theta}(\mu_{A}(x))$&$\lambda_{e}$\\
$a_{r}$ & $m_{\theta}(\mu_{A}(x))\leq \beta$& $0$ & $m_{\theta}(\mu_{A}(x))$&$\lambda_{r}$\\
$a_{s_{\downarrow}}$ & $0.5\leq m_{\theta}(\mu_{A}(x))<\alpha$& $0.5$ &$m_{\theta}(\mu_{A}(x))-0.5$ &$\lambda_{s_{\downarrow}}$ \\
$a_{s_{\uparrow}}$ & $\beta<m_{\theta}(\mu_{A}(x))<0.5$& $0.5$ &$0.5-m_{\theta}(\mu_{A}(x))$ & $\lambda_{s_{\uparrow}}$\\
\hline
\end{tabular}
\end{table}

Table $2$ summarizes information about three-way approximations of an
interval-valued fuzzy set. Concretely, the set of actions
$\{a_{e},a_{r}, a_{s_{\downarrow}},a_{s_{\uparrow}}\}$ describes
four possible actions on changing the membership grade. For
simplicity, we also use $\{e,r, s_{\downarrow},s_{\uparrow}\}$ to
denote the four actions. The elevation action $a_{e}$
elevate the membership grade of $x$ from $m_{\theta}(\mu_{A}(x))$ to
$1$, the reduction action $a_{r}$ reduce the membership grade of $x$
from $m_{\theta}(\mu_{A}(x))$ to $0$, the elevation
$a_{s_{\uparrow}}$ elevate the membership grade of $x$ from
$m_{\theta}(\mu_{A}(x))$ to $0.5$ if $m_{\theta}(\mu_{A}(x))<0.5$,
the reduction $a_{s_{\downarrow}}$ reduce the membership grade of
$x$ from $m_{\theta}(\mu_{A}(x))$ to $0.5$ if
$m_{\theta}(\mu_{A}(x))>0.5$. The fuzzy membership grade
$m_{\theta}(\mu_{A}(x))$ represents the state of object in the
second column, and the errors of different actions are given in the
fourth column, and the losses of different actions are given in the
fifth column.

Each of the four losses $\lambda_{e},\lambda_{r},
\lambda_{s_{\downarrow}}$ and $\lambda_{s_{\uparrow}}$ provides the unit
cost, and the actual cost of each action is weighted by the
magnitude of its error. Suppose
$R_{a}(x)=\lambda_{a}E_{a}(\mu_{A}(x))$ denote the loss for
taking actions $\{e,r, s_{\downarrow},s_{\uparrow}\}$, the losses of four
actions for an object can be computed as
\begin{eqnarray*}
R_{e}(x)&=&\lambda_{e}E_{e}(\mu_{A}(x))=(1-m_{\theta}(\mu_{A}(x)))\lambda_{e};\\
R_{r}(x)&=&\lambda_{r}E_{r}(\mu_{A}(x))=m_{\theta}(\mu_{A}(x))\lambda_{r};
\\
R_{s_{\downarrow}}(x)&=&\lambda_{s_{\downarrow}}E_{s_{\downarrow}}(m_{\theta}(\mu_{A}(x)))=(m_{\theta}(\mu_{A}(x))-0.5)\lambda_{s_{\downarrow}};\\
R_{s_{\uparrow}}(x)&=&\lambda_{s_{\uparrow}}E_{s_{\uparrow}}(m_{\theta}(\mu_{A}(x)))=(0.5-m_{\theta}(\mu_{A}(x)))\lambda_{s_{\uparrow}}.
\end{eqnarray*}

Since only an action is taken for each object, the total loss of the
approximation is computed by
\begin{eqnarray*}
R=\sum_{x\in
U}R_{a}(x)=\sum_{x\in U}\lambda_{a}E_{a}(\mu_{A}(x)).
\end{eqnarray*}

To minimize the total loss $R$, we take an action $\tau(x)$
that minimizes the loss $R_{a}(x)$ for each object, and
$\tau(x)$ is a solution to the following minimization problem as
$$
\text{arg } min_{a\in action}R_{a}(x),
$$
where $a\in \{e,r, s_{\downarrow},s_{\uparrow}\}$.

According to the value $\mu_{A}(x)$ of an object $x$, we have two
groups of decision rules for obtaining three-way approximations of an
interval-valued fuzzy set as follows:

$(1)$ When $m_{\theta}(\mu_{A}(x))\geq0.5$, $(E1)$ If $R(a_{e}|x)\leq
R(a_{r}|x)$ and $R(a_{e}|x)\leq R(a_{s_{\downarrow}}|x)$, then take
action $a_{e}$; $(R1)$ If $R(a_{r}|x)\leq R(a_{e}|x)$ and
$R(a_{r}|x)\leq R(a_{s_{\downarrow}}|x)$, then take action $a_{r}$;
$(S1)$ If $R(a_{s_{\downarrow}}|x)\leq R(a_{e}|x)$ and
$R(a_{s_{\downarrow}}|x)\leq R(a_{r}|x)$, then take action
$a_{s_{\downarrow}}$.

$(2)$ When $m_{\theta}(\mu_{A}(x))<0.5$, $(E2)$ If $R(a_{e}|x)\leq R(a_{r}|x)$
and $R(a_{e}|x)\leq R(a_{s_{\uparrow}}|x)$, then take action
$a_{e}$; $(R2)$ If $R(a_{r}|x)\leq R(a_{e}|x)$ and $R(a_{r}|x)\leq
R(a_{s_{\uparrow}}|x)$, then take action $a_{r}$; $(S2):$ If
$R(a_{s_{\uparrow}}|x)\leq R(a_{e}|x)$ and
$R(a_{s_{\uparrow}}|x)\leq R(a_{r}|x)$, then take action
$a_{s_{\uparrow}}$.

\subsection{Single-valued loss functions-based three-way approximations of interval-valued
fuzzy sets}

In this subsection, we consider loss functions satisfying certain
properties for obtaining an analytic solution defining a three-way
approximation.

Suppose $(c1):
\lambda_{e}>0,\lambda_{r}>0,\lambda_{s_{\downarrow}}>0,\lambda_{s_{\uparrow}}>0;
$ $(c2): \lambda_{s_{\downarrow}}\leq\lambda_{r};$ $(c3):
\lambda_{s_{\uparrow}}\leq\lambda_{e}$, Condition $(c1)$ requires
that all costs are nonnegative; Condition $(c2)$ illustrates that
reducing a membership grade $\mu_{A}(x)\geq 0.5$ to $0.5$ represents
a smaller adjustment than reducing it to $0$, and a smaller cost is
associated with action $a_{s_{\downarrow}}$; Condition $(c3)$
illustrates that elevating a membership grade $\mu_{A}(x)< 0.5$ to
$0.5$ represents a smaller adjustment than elevating it to $1$, and
a smaller cost is associated with action $a_{s_{\uparrow}}$. With
the assumptions $(c1)-(c3)$, we simplify the decision rules as
follows:

$(1)$ When $m_{\theta}(\mu_{A}(x))\geq 0.5$, the rule $(E1)$ is
expressed as
\begin{eqnarray*}
R(a_{e}|x)\leq R(a_{r}|x)&\Leftrightarrow&
(1-m_{\theta}(\mu_{A}(x)))\lambda_{e}\leq
m_{\theta}(\mu_{A}(x))\lambda_{r}\\&\Leftrightarrow&
\mu_{A}(x)\geq
\frac{\lambda_{e}}{\lambda_{e}+\lambda_{r}}=\gamma;\\
R(a_{e}|x)\leq R(a_{s_{\downarrow}}|x)&\Leftrightarrow&
(1-m_{\theta}(\mu_{A}(x)))\lambda_{e}\leq
(m_{\theta}(\mu_{A}(x))-0.5)\lambda_{s_{\downarrow}}\\&\Leftrightarrow&
m_{\theta}(\mu_{A}(x))\geq
\frac{2\lambda_{e}+\lambda_{s_{\downarrow}}}{2(\lambda_{e}+\lambda_{s_{\downarrow}})}=\alpha.
\end{eqnarray*}

The rule $R(1)$ is expressed by
\begin{eqnarray*}
R(a_{r}|x)\leq R(a_{r}|x)&\Leftrightarrow&
m_{\theta}(\mu_{A}(x))\leq \gamma
;\\
R(a_{r}|x)\leq R(a_{s_{\downarrow}}|x)&\Leftrightarrow&
m_{\theta}(\mu_{A}(x))\lambda_{r}\leq
(m_{\theta}(\mu_{A}(x))-0.5)\lambda_{s_{\downarrow}}\\&\Leftrightarrow&
m_{\theta}(\mu_{A}(x))\leq
\frac{-\lambda_{s_{\downarrow}}(t)}{2(\lambda_{r}-\lambda_{s_{\downarrow}}(t))}=\gamma^{-}.
\end{eqnarray*}

The rule $S(1)$ is expressed by
\begin{eqnarray*}
R(a_{s_{\downarrow}}|x)\leq R(a_{e}|x)\Leftrightarrow
m_{\theta}(\mu_{A}(x))\leq\alpha
;
R(a_{s_{\downarrow}}|x)\leq R(a_{s_{\downarrow}}|x)\Leftrightarrow
m_{\theta}(\mu_{A}(x))\geq\gamma^{-}.
\end{eqnarray*}

Since $\gamma^{-}\leq 0$ contradicts with the assumption
$m_{\theta}(\mu_{A}(x))\geq 0.5$, it is impossible to apply rule
$(R1)$ for reducing membership values. Therefore, when
$m_{\theta}(\mu_{A}(x))\geq 0.5$, the rules are simply expressed as
(E1) If $m_{\theta}(\mu_{A}(x))\geq \alpha$, then
$T_{\mu_{A}}(x)=1$;
(S1) If $0.5\leq m_{\theta}(\mu_{A}(x))< \alpha$, then
$T_{\mu_{A}}(x)=0.5$.

$(2)$ When $m_{\theta}(\mu_{A}(x))< 0.5$, the rule $(E2)$ is
expressed as
\begin{eqnarray*}
R(a_{e}|x)\leq R(a_{r}|x)&\Leftrightarrow&
(1-m_{\theta}(\mu_{A}(x)))\lambda_{e}\leq
m_{\theta}(\mu_{A}(x))\lambda_{r}\\&\Leftrightarrow&
m_{\theta}(\mu_{A}(x))\geq
\frac{\lambda_{e}}{\lambda_{e}+\lambda_{r}}=\gamma;\\
R(a_{e}|x)\leq R(a_{s_{\uparrow}}|x)&\Leftrightarrow&
(1-m_{\theta}(\mu_{A}(x)))\lambda_{e}\leq
(0.5-m_{\theta}(\mu_{A}(x)))\lambda_{s_{\uparrow}}\\&\Leftrightarrow&
m_{\theta}(\mu_{A}(x))\geq
\frac{\lambda_{e}-0.5\lambda_{s_{\uparrow}}}{\lambda_{e}-\lambda_{s_{\uparrow}}}=\gamma^{+}.
\end{eqnarray*}

The rule $R(2)$ is expressed as
\begin{eqnarray*}
R(a_{r}|x)\leq R(a_{e}|x)&\Leftrightarrow&
m_{\theta}(\mu_{A}(x))\leq \gamma
;\\
R(a_{r}|x)\leq R(a_{s_{\uparrow}}|x)&\Leftrightarrow&
m_{\theta}(\mu_{A}(x))\lambda_{r}\leq
(0.5-m_{\theta}(\mu_{A}(x)))\lambda_{s_{\uparrow}}\\&\Leftrightarrow&
m_{\theta}(\mu_{A}(x))\leq
\frac{\lambda_{s_{\uparrow}}}{2(\lambda_{r}+\lambda_{s_{\uparrow}})}=\beta.
\end{eqnarray*}

The rule $S(2)$ is expressed as
\begin{eqnarray*}
R(a_{s_{\uparrow}}|x)\leq R(a_{e}|x)\Leftrightarrow
m_{\theta}(\mu_{A}(x))\leq\gamma^{+}
;
R(a_{s_{\uparrow}}|x)\leq R(a_{s_{\uparrow}}|x)\Leftrightarrow
m_{\theta}(\mu_{A}(x))\geq\beta.
\end{eqnarray*}

Since $\gamma^{+}\geq 1$ contradicts with the assumption
$m_{\theta}(\mu_{A}(x))< 0.5$, it is impossible to apply rule $(E2)$
for elevating membership values. Therefore, when
$m_{\theta}(\mu_{A}(x))<0.5$, the remaining rules are simply
expressed as
(R2) If $m_{\theta}(\mu_{A}(x))\leq\beta$, then $T_{\mu_{A}}(x)=0$;
(S2) If $\beta\leq m_{\theta}(\mu_{A}(x))< 0.5$, then
$T_{\mu_{A}}(x)=0.5$.

By combining the two sets of rules, we immediately have three rules
as (E) If $m_{\theta}(\mu_{A}(x))\geq \alpha(t)$, then
$T_{\mu_{A}}(x)=1$; (R) If $m_{\theta}(\mu_{A}(x))\leq\beta$, then
$T_{\mu_{A}}(x)=0$; (S) If $\beta<m_{\theta}(\mu_{A}(x))<\alpha$,
then $T_{\mu_{A}}(x)=0.5$, where
\begin{eqnarray*}
\alpha=\frac{2\lambda_{e}+\lambda_{s_{\downarrow}}}{2(\lambda_{e}(t)+\lambda_{s_{\downarrow}})}
\text{ and
}\beta=\frac{\lambda_{s_{\uparrow}}}{2(\lambda_{r}+\lambda_{s_{\uparrow}})}.
\end{eqnarray*}

\section{Interval-valued loss functions-based three-way approximations
of interval-valued fuzzy sets: I}

In this section, we introduce a framework for interval-valued loss functions-based three-way approximations of interval-valued fuzzy sets.

\subsection{Cost-sensitive three-way approximations of interval-valued fuzzy sets}

In Section 4, we investigate three-way approximation of interval-valued
fuzzy sets by using three membership grades of $0$, $0.5$ and $1$.
We take one of the following three actions for an object with a
membership grade: elevate the membership grade to 1, reduce the
membership grade to 0, and change the membership grade to 0.5. More
specially, there are two situations for the third case: reduce the
membership grade to 0.5 if $m_{\theta}(\mu_{A}(x))\geq 0.5$ and
elevate the membership grade to 0.5 if $m_{\theta}(\mu_{A}(x))<
0.5$. Each action will incur error and the costs of different
actions are not necessarily the same.

\begin{table}[htbp]\renewcommand{\arraystretch}{1.5}
\caption{Interval-valued loss function.}
 \tabcolsep0.07in
\begin{tabular}{c c c c c}
\hline  Action & \text{Membership grade} &\text{Three-way membership grade}& Error &Loss\\
\hline
$a_{e}$ & $m_{\theta}(\mu_{A}(x))\geq \alpha$& $1$  &$1- m_{\theta}(\mu_{A}(x))$&$\widetilde{\lambda}_{e}=[\lambda^{-}_{e},\lambda^{+}_{e}]$\\
$a_{r}$ & $m_{\theta}(\mu_{A}(x))\leq \beta$& $0$ & $m_{\theta}(\mu_{A}(x))$&$\widetilde{\lambda}_{r}=[\lambda^{-}_{r},\lambda^{+}_{r}]$\\
$a_{s_{\downarrow}}$ & $0.5\leq m_{\theta}(\mu_{A}(x))<\alpha$& $0.5$ &$m_{\theta}(\mu_{A}(x))-0.5$ &$\widetilde{\lambda}_{s_{\downarrow}}=[\lambda^{-}_{s_{\downarrow}},\lambda^{+}_{s_{\downarrow}}]$ \\
$a_{s_{\uparrow}}$ & $\beta<m_{\theta}(\mu_{A}(x))<0.5$& $0.5$ &$0.5-m_{\theta}(\mu_{A}(x))$ & $\widetilde{\lambda}_{s\uparrow}=[\lambda^{-}_{s\uparrow},\lambda^{+}_{s\uparrow}]$\\
\hline
\end{tabular}
\end{table}

Table $3$ summarizes information about three-way approximations of an
interval-valued fuzzy set. Concretely, the set of actions
$\{a_{e},a_{r}, a_{s_{\downarrow}},a_{s_{\uparrow}}\}$ describes
four possible actions on changing the membership grade. For
simplicity, we also use $\{e,r, s_{\downarrow},s_{\uparrow}\}$ to
denote the four actions. Concretely, the elevation action $a_{e}$
elevate the membership grade of $x$ from $m_{\theta}(\mu_{A}(x))$ to
$1$, the reduction action $a_{r}$ reduce the membership grade of $x$
from $m_{\theta}(\mu_{A}(x))$ to $0$, the elevation
$a_{s_{\uparrow}}$ elevate the membership grade of $x$ from
$m_{\theta}(\mu_{A}(x))$ to $0.5$ if $m_{\theta}(\mu_{A}(x))<0.5$,
the reduction $a_{s_{\downarrow}}$ reduce the membership grade of
$x$ from $m_{\theta}(\mu_{A}(x))$ to $0.5$ if
$m_{\theta}(\mu_{A}(x))>0.5$. The fuzzy membership grade
$m_{\theta}(\mu_{A}(x))$ represents the state of object in the
second column, and the errors of different actions are given in the
fourth column, and the losses of different actions are given in the
fifth column.

Each of the four losses
$m_{\theta}(\widetilde{\lambda}_{e}),m_{\theta}(\widetilde{\lambda}_{r}),
m_{\theta}(\widetilde{\lambda}_{s_{\downarrow}})$ and
$m_{\theta}(\widetilde{\lambda}_{s_{\uparrow}})$ provides the unit
cost, and the actual cost of each action is weighted by the
magnitude of its error. Suppose
$R_{a}(x)=m_{\theta}(\widetilde{\lambda}_{a})E_{a}(\mu_{A}(x))$
denote the loss for taking actions $\{e,r,
s_{\downarrow},s_{\uparrow}\}$, the losses of four actions for an
object can be computed as
\begin{eqnarray*}
R_{e}(x)&=&m_{\theta}(\widetilde{\lambda}_{e})E_{e}(\mu_{A}(x))=(1-m_{\theta}(\mu_{A}(x)))m_{\theta}(\widetilde{\lambda}_{e});\\
R_{r}(x)&=&m_{\theta}(\widetilde{\lambda}_{r})E_{r}(\mu_{A}(x))=m_{\theta}(\mu_{A}(x))m_{\theta}(\widetilde{\lambda}_{r});
\\
R_{s_{\downarrow}}(x)&=&m_{\theta}(\widetilde{\lambda}_{s_{\downarrow}})E_{s_{\downarrow}}(m_{\theta}(\mu_{A}(x)))=(m_{\theta}(\mu_{A}(x))-0.5)m_{\theta}(\widetilde{\lambda}_{s_{\downarrow}});\\
R_{s_{\uparrow}}(x)&=&m_{\theta}(\widetilde{\lambda}_{s_{\uparrow}})E_{s_{\uparrow}}(m_{\theta}(\mu_{A}(x)))=(0.5-m_{\theta}(\mu_{A}(x)))m_{\theta}(\widetilde{\lambda}_{s_{\uparrow}}).
\end{eqnarray*}

Since only an action is taken for each object, the total loss of the
approximation is computed by
\begin{eqnarray*}
R=\sum_{x\in U}R_{a}(x)=\sum_{x\in
U}m_{\theta}(\widetilde{\lambda}_{a})E_{a}(\mu_{A}(x)).
\end{eqnarray*}

To minimize the total loss $R$, we take an action $\tau(x)$ that
minimizes the loss $R_{a}(x)$ for each object, and $\tau(x)$ is a
solution to the following minimization problem as
$$
\text{arg } min_{a\in action}R_{a}(x),
$$
where $a\in \{e,r, s_{\downarrow},s_{\uparrow}\}$.

According to the value $\mu_{A}(x)$ of an object $x$, we have two
groups of decision rules for obtaining three-way approximations of
an interval-valued fuzzy set as follows:

$(1)$ When $m_{\theta}(\mu_{A}(x))\geq0.5$, $(E1)$ If
$R(a_{e}|x)\leq R(a_{r}|x)$ and $R(a_{e}|x)\leq
R(a_{s_{\downarrow}}|x)$, then take action $a_{e}$; $(R1)$ If
$R(a_{r}|x)\leq R(a_{e}|x)$ and $R(a_{r}|x)\leq
R(a_{s_{\downarrow}}|x)$, then take action $a_{r}$; $(S1)$ If
$R(a_{s_{\downarrow}}|x)\leq R(a_{e}|x)$ and
$R(a_{s_{\downarrow}}|x)\leq R(a_{r}|x)$, then take action
$a_{s_{\downarrow}}$.

$(2)$ When $m_{\theta}(\mu_{A}(x))<0.5$, $(E2)$ If $R(a_{e}|x)\leq
R(a_{r}|x)$ and $R(a_{e}|x)\leq R(a_{s_{\uparrow}}|x)$, then take
action $a_{e}$; $(R2)$ If $R(a_{r}|x)\leq R(a_{e}|x)$ and
$R(a_{r}|x)\leq R(a_{s_{\uparrow}}|x)$, then take action $a_{r}$;
$(S2):$ If $R(a_{s_{\uparrow}}|x)\leq R(a_{e}|x)$ and
$R(a_{s_{\uparrow}}|x)\leq R(a_{r}|x)$, then take action
$a_{s_{\uparrow}}$.

\subsection{Loss functions-based three-way approximations of interval-valued
fuzzy sets}

In this subsection, we consider interval-valued loss functions satisfying
certain properties for obtaining an analytic solution defining a
three-way approximation.

Suppose $(c1):
m_{\theta}(\widetilde{\lambda}_{e})>0,m_{\theta}(\widetilde{\lambda}_{r})>0,m_{\theta}(\widetilde{\lambda}_{s_{\downarrow}})>0,m_{\theta}(\widetilde{\lambda}_{s_{\uparrow}})>0;
$ $(c2): m_{\theta}(\widetilde{\lambda}_{s_{\downarrow}})\leq m_{\theta}(\widetilde{\lambda}_{r});$ $(c3):
m_{\theta}(\widetilde{\lambda}_{s_{\uparrow}})\leq m_{\theta}(\widetilde{\lambda}_{e})$, Condition $(c1)$ requires
that all costs are nonnegative; Condition $(c2)$ illustrates that
reducing a membership grade $\mu_{A}(x)\geq 0.5$ to $0.5$ represents
a smaller adjustment than reducing it to $0$, and a smaller cost is
associated with action $a_{s_{\downarrow}}$; Condition $(c3)$
illustrates that elevating a membership grade $\mu_{A}(x)< 0.5$ to
$0.5$ represents a smaller adjustment than elevating it to $1$, and
a smaller cost is associated with action $a_{s_{\uparrow}}$. With
the assumptions $(c1)-(c3)$, we simplify the decision rules as
follows:

$(1)$ When $m_{\theta}(\mu_{A}(x))\geq 0.5$, the rule $(E1)$ is
expressed as
\begin{eqnarray*}
R(a_{e}|x)\leq R(a_{r}|x)&\Leftrightarrow&
(1-m_{\theta}(\mu_{A}(x)))m_{\theta}(\widetilde{\lambda}_{e})\leq
(m_{\theta}(\mu_{A}(x))-0)m_{\theta}(\widetilde{\lambda}_{r})\\&\Leftrightarrow&
\mu_{A}(x)\geq
\frac{m_{\theta}(\widetilde{\lambda}_{e})}{\lambda_{e}m_{\theta}(\widetilde{\lambda}_{e})+m_{\theta}(\widetilde{\lambda}_{r})}=\gamma;\\
R(a_{e}|x)\leq R(a_{s_{\downarrow}}|x)&\Leftrightarrow&
(1-m_{\theta}(\mu_{A}(x)))m_{\theta}(\widetilde{\lambda}_{e})\leq
(m_{\theta}(\mu_{A}(x))-0.5)m_{\theta}(\widetilde{\lambda}_{s_{\downarrow}})\\&\Leftrightarrow&
m_{\theta}(\mu_{A}(x))\geq
\frac{2m_{\theta}(\widetilde{\lambda}_{e})+m_{\theta}(\widetilde{\lambda}_{s_{\downarrow}})}{2(m_{\theta}(\widetilde{\lambda}_{e})+m_{\theta}(\widetilde{\lambda}_{s_{\downarrow}})}=\alpha.
\end{eqnarray*}

The rule $R(1)$ is expressed by
\begin{eqnarray*}
R(a_{r}|x)\leq R(a_{r}|x)&\Leftrightarrow&
m_{\theta}(\mu_{A}(x))\leq \gamma
;\\
R(a_{r}|x)\leq R(a_{s_{\downarrow}}|x)&\Leftrightarrow&
m_{\theta}(\mu_{A}(x))\lambda_{r}\leq
(m_{\theta}(\mu_{A}(x))-0.5)m_{\theta}(\widetilde{\lambda}_{s_{\downarrow}})\\&\Leftrightarrow&
m_{\theta}(\mu_{A}(x))\leq
\frac{-m_{\theta}(\widetilde{\lambda}_{s_{\downarrow}})}{2(m_{\theta}(\widetilde{\lambda}_{r})-m_{\theta}(\widetilde{\lambda}_{s_{\downarrow}})(t))}=\gamma^{-}.
\end{eqnarray*}

The rule $S(1)$ is expressed by
\begin{eqnarray*}
R(a_{s_{\downarrow}}|x)\leq R(a_{e}|x)\Leftrightarrow
m_{\theta}(\mu_{A}(x))\leq\alpha
;
R(a_{s_{\downarrow}}|x)\leq R(a_{s_{\downarrow}}|x)\Leftrightarrow
m_{\theta}(\mu_{A}(x))\geq\gamma^{-}.
\end{eqnarray*}

Since $\gamma^{-}\leq 0$ contradicts with the assumption
$m_{\theta}(\mu_{A}(x))\geq 0.5$, it is impossible to apply rule
$(R1)$ for reducing membership values. Therefore, when
$m_{\theta}(\mu_{A}(x))\geq 0.5$, the rules are simply expressed as
(E1) If $m_{\theta}(\mu_{A}(x))\geq \alpha$, then
$T_{\mu_{A}}(x)=1$;
(S1) If $0.5\leq m_{\theta}(\mu_{A}(x))< \alpha$, then
$T_{\mu_{A}}(x)=0.5$.

$(2)$ When $m_{\theta}(\mu_{A}(x))< 0.5$, the rule $(E2)$ is
expressed as
\begin{eqnarray*}
R(a_{e}|x)\leq R(a_{r}|x)&\Leftrightarrow&
(1-m_{\theta}(\mu_{A}(x)))m_{\theta}(\widetilde{\lambda}_{e})\leq
(m_{\theta}(\mu_{A}(x))-0)m_{\theta}(\widetilde{\lambda}_{r})\\&\Leftrightarrow&
m_{\theta}(\mu_{A}(x))\geq
\frac{m_{\theta}(\widetilde{\lambda}_{e})}{m_{\theta}(\widetilde{\lambda}_{e})+m_{\theta}(\widetilde{\lambda}_{r})}=\gamma;\\
R(a_{e}|x)\leq R(a_{s_{\uparrow}}|x)&\Leftrightarrow&
(1-m_{\theta}(\mu_{A}(x)))m_{\theta}(\widetilde{\lambda}_{e})\leq
(0.5-m_{\theta}(\mu_{A}(x)))m_{\theta}(\widetilde{\lambda}_{s_{\uparrow}})\\&\Leftrightarrow&
m_{\theta}(\mu_{A}(x))\geq
\frac{m_{\theta}(\widetilde{\lambda}_{e})-0.5m_{\theta}(\widetilde{\lambda}_{s_{\uparrow}})}{m_{\theta}(\widetilde{\lambda}_{e})-m_{\theta}(\widetilde{\lambda}_{s_{\uparrow}})}=\gamma^{+}.
\end{eqnarray*}

The rule $R(2)$ is expressed as
\begin{eqnarray*}
R(a_{r}|x)\leq R(a_{e}|x)&\Leftrightarrow&
m_{\theta}(\mu_{A}(x))\leq \gamma
;\\
R(a_{r}|x)\leq R(a_{s_{\uparrow}}|x)&\Leftrightarrow&
m_{\theta}(\mu_{A}(x))m_{\theta}(\widetilde{\lambda}_{r})\leq
(0.5-m_{\theta}(\mu_{A}(x)))m_{\theta}(\widetilde{\lambda}_{s_{\uparrow}})\\&\Leftrightarrow&
m_{\theta}(\mu_{A}(x))\leq
\frac{m_{\theta}(\widetilde{\lambda}_{s_{\uparrow}})}{2(m_{\theta}(\widetilde{\lambda}_{r})+m_{\theta}(\widetilde{\lambda}_{s_{\uparrow}}))}=\beta.
\end{eqnarray*}

The rule $S(2)$ is expressed as
\begin{eqnarray*}
R(a_{s_{\uparrow}}|x)\leq R(a_{e}|x)\Leftrightarrow
m_{\theta}(\mu_{A}(x))\leq\gamma^{+}
;
R(a_{s_{\uparrow}}|x)\leq R(a_{s_{\uparrow}}|x)\Leftrightarrow
m_{\theta}(\mu_{A}(x))\geq\beta.
\end{eqnarray*}

Since $\gamma^{+}\geq 1$ contradicts with the assumption
$m_{\theta}(\mu_{A}(x))< 0.5$, it is impossible to apply rule $(E2)$
for elevating membership values. Therefore, when
$m_{\theta}(\mu_{A}(x))<0.5$, the remaining rules are simply
expressed as
(R2) If $m_{\theta}(\mu_{A}(x))\leq\beta$, then $T_{\mu_{A}}(x)=0$;
(S2) If $\beta\leq m_{\theta}(\mu_{A}(x))< 0.5$, then
$T_{\mu_{A}}(x)=0.5$.

By combining the two sets of rules, we immediately have three rules
as (E) If $m_{\theta}(\mu_{A}(x))\geq \alpha(t)$, then
$T_{\mu_{A}}(x)=1$; (R) If $m_{\theta}(\mu_{A}(x))\leq\beta$, then
$T_{\mu_{A}}(x)=0$; (S) If $\beta<m_{\theta}(\mu_{A}(x))<\alpha$,
then $T_{\mu_{A}}(x)=0.5$, where
\begin{eqnarray*}
\alpha=\frac{2m_{\theta}(\widetilde{\lambda}_{e})+m_{\theta}(\widetilde{\lambda}_{s_{\downarrow}})}{2(m_{\theta}(\widetilde{\lambda}_{e})+m_{\theta}(\widetilde{\lambda}_{s_{\downarrow}}))}
\text{ and
}\beta=\frac{m_{\theta}(\widetilde{\lambda}_{s_{\uparrow}})}{2(m_{\theta}(\widetilde{\lambda}_{r})+m_{\theta}(\widetilde{\lambda}_{s_{\uparrow}}))}.
\end{eqnarray*}

\section{Interval-valued loss functions-based three-way approximations
of interval-valued fuzzy sets: II}

In this section, we introduce another framework for decision-theoretic
rough sets-based three-way approximations of interval-valued fuzzy sets.

\begin{definition}
Let $\widetilde{\mu}_{1}=[\mu^{-}_{1},\mu^{+}_{1}]$ and $\widetilde{\mu}_{2}=[\lambda^{-}_{2},\mu^{+}_{2}]$ be interval-valued
sets, then the degree of possibility of $\widetilde{\mu}_{1}\geq \widetilde{\mu}_{2}$ and $\widetilde{\mu}_{2}\geq \widetilde{\mu}_{1}$ are defined as
\begin{eqnarray*}
p(\widetilde{\mu}_{1}\geq \widetilde{\mu}_{2})&=&max\{1-max\{\frac{\mu_{2}^{+}-\mu_{1}^{-}}{\mu_{1}^{+}-\mu_{1}^{-}+\mu_{2}^{+}-\mu_{2}^{-}},0\},0\};\\
p(\widetilde{\mu}_{2}\geq \widetilde{\mu}_{1})&=&max\{1-max\{\frac{\mu_{1}^{+}-\mu_{2}^{-}}{\mu_{1}^{+}-\mu_{1}^{-}+\mu_{2}^{+}-\mu_{2}^{-}},0\},0\}.
\end{eqnarray*}
\end{definition}

In the sense of Definition 6.1, we have
\makeatother $$p(\widetilde{\mu}_{1}\geq \widetilde{\mu}_{2})=\left\{
\begin{array}{ccc}
0,&{\rm }& \frac{\mu_{2}^{+}-\mu_{1}^{-}}{\mu_{1}^{+}-\mu_{1}^{-}+\mu_{2}^{+}-\mu_{2}^{-}}\geq1;\\
1-\frac{\mu_{2}^{+}-\mu_{1}^{-}}{\mu_{1}^{+}-\mu_{1}^{-}+\mu_{2}^{+}-\mu_{2}^{-}},&{\rm }& 0<\frac{\mu_{2}^{+}-\mu_{1}^{-}}{\mu_{1}^{+}-\mu_{1}^{-}+\mu_{2}^{+}-\mu_{2}^{-}}<1;\\
1,&{\rm }& \frac{\mu_{2}^{+}-\mu_{1}^{-}}{\mu_{1}^{+}-\mu_{1}^{-}+\mu_{2}^{+}-\mu_{2}^{-}}\leq0.
\end{array}
\right. $$
Furthermore, we have the  complementary matrix of the
preference as
\begin{eqnarray*}
P_{\widetilde{\mu}_{1}\widetilde{\mu}_{2}\widetilde{\mu}_{3}}=\left[
\begin{array}{ccc}
p(\widetilde{\mu}_{1}\geq \widetilde{\mu}_{1})  & p(\widetilde{\mu}_{1}\geq \widetilde{\mu}_{2})  & p(\widetilde{\mu}_{1}\geq \widetilde{\mu}_{3}) \\
p(\widetilde{\mu}_{2}\geq \widetilde{\mu}_{1})  & p(\widetilde{\mu}_{2}\geq \widetilde{\mu}_{2})  & p(\widetilde{\mu}_{2}\geq \widetilde{\mu}_{3})\\
p(\widetilde{\mu}_{3}\geq \widetilde{\mu}_{1})  & p(\widetilde{\mu}_{3}\geq \widetilde{\mu}_{2})  & p(\widetilde{\mu}_{3}\geq \widetilde{\mu}_{3})\\
\end{array}
\right].
\end{eqnarray*}

Suppose
$\widetilde{R}_{a}(x)=\widetilde{\lambda}_{a}E_{a}(\mu_{A}(x))$ denote the loss for
taking actions $\{e,r, s_{\downarrow},s_{\uparrow}\}$, the losses of four
actions for an object can be computed as
\begin{eqnarray*}
\widetilde{R}_{e}(x)&=&\widetilde{\lambda}_{e}E_{e}(\mu_{A}(x))=[(1-m_{\theta}(\mu_{A}(x)))\lambda^{-}_{e},(1-m_{\theta}(\mu_{A}(x)))\lambda^{+}_{e}];\\
\widetilde{R}_{r}(x)&=&\widetilde{\lambda}_{r}E_{r}(\mu_{A}(x))=[m_{\theta}(\mu_{A}(x))\lambda^{-}_{r},m_{\theta}(\mu_{A}(x))\lambda^{+}_{r}];
\\
\widetilde{R}_{s_{\downarrow}}(x)&=&\widetilde{\lambda}_{s_{\downarrow}}E_{s_{\downarrow}}(m_{\theta}(\mu_{A}(x)))=[(m_{\theta}(\mu_{A}(x))-0.5)\lambda^{-}_{s_{\downarrow}},(m_{\theta}(\mu_{A}(x))-0.5)\lambda^{+}_{s_{\downarrow}}];\\
\widetilde{R}_{s_{\uparrow}}(x)&=&\widetilde{\lambda}_{s_{\uparrow}}E_{s_{\uparrow}}(m_{\theta}(\mu_{A}(x)))=[(0.5-m_{\theta}(\mu_{A}(x)))\lambda^{-}_{s_{\uparrow}},(0.5-m_{\theta}(\mu_{A}(x)))\lambda^{+}_{s_{\uparrow}}].
\end{eqnarray*}

Since only an action is taken for each object, the total loss of the
approximation is computed by
\begin{eqnarray*}
\widetilde{R}=\sum_{x\in
U}\widetilde{R}_{a}(x)=\sum_{x\in U}\widetilde{\lambda}_{a}E_{a}(m_{\theta}(\mu_{A}(x))).
\end{eqnarray*}

To minimize the total loss $\widetilde{R}$, we take an action $\tau(x)$
that minimizes the loss $\widetilde{R}_{a}(x)$ for each object, and
$\tau(x)$ is a solution to the following minimization problem as
$$
\text{arg } min_{a\in action}\widetilde{R}_{a}(x),
$$
where $a\in \{e,r, s_{\downarrow},s_{\uparrow}\}$.

According to the value $\mu_{A}(x)$ of an object $x$, we have two
groups of decision rules for obtaining three-way approximations of an
interval-valued fuzzy set.

In what follows, in light of complementary matrix of the
preference, we discuss the ranking of the expected loss
$\widetilde{R}_{a}(x)$ and generate decision rules in the context of
interval-valued fuzzy sets. Concretely, there are two situations to
discuss: $m_{\theta}(\mu_{A}(x))\geq 0.5$ and
$m_{\theta}(\mu_{A}(x))< 0.5$.

\subsection{Situation 1: $m_{\theta}(\mu_{A}(x))\geq 0.5$}

If $m_{\theta}(\mu_{A}(x))\geq 0.5$, then we have the complementary
matrix of the preference as
\begin{eqnarray*}
P_{ers_{\downarrow}}=\left[
\begin{array}{ccc}
p_{ee}  & p_{er}  & p_{es_{\downarrow}} \\
p_{re}  & p_{rr}  & p_{rs_{\downarrow}}\\
p_{s_{\downarrow}e}  & p_{s_{\downarrow}r}  & p_{s_{\downarrow}s_{\downarrow}}\\
\end{array}
\right].
\end{eqnarray*}

According to the properties of the degree of possibilities, we have
$p_{ee}=p_{rr}=p_{s_{\downarrow}s_{\downarrow}}=0.5,p_{er}+p_{re}=1,
p_{es_{\downarrow}}+p_{s_{\downarrow}e}=1$ and $ p_{rs_{\downarrow}}+p_{s_{\downarrow}r}=1.$ Then we simplify the complementary matrix as
\begin{eqnarray*}
P_{ers_{\downarrow}}=\left[
\begin{array}{ccc}
0.5  & p_{er}  & p_{es_{\downarrow}} \\
1-p_{er}  & 0.5  & p_{rs_{\downarrow}}\\
1-p_{es_{\downarrow}}  & 1-p_{rs_{\downarrow}}  & 0.5\\
\end{array}
\right].
\end{eqnarray*}

In light of the complementary matrix $P_{ers_{\downarrow}}$, all elements in each
line of the matrix are summarized as
\begin{eqnarray*}
p_{e}=0.5+p_{er}+ p_{es_{\downarrow}};
p_{r}=0.5-p_{er}+p_{rs_{\downarrow}};
p_{s_{\downarrow}}=2.5-p_{es_{\downarrow}}-p_{rs_{\downarrow}},
\end{eqnarray*}
where $p_{e}$ is the total degree of preference of $\widetilde{R}_{e}(x);$
$p_{r}$ is the total degree of preference of $\widetilde{R}_{r}(x);$
$p_{s_{\downarrow}}$ is the total degree of preference of $\widetilde{R}_{s_{\downarrow}}(x)$. The values of $p_{e},p_{r}$ and $p_{s_{\downarrow}}$ depend on $p_{er},p_{es_{\downarrow}}$ and $p_{rs_{\downarrow}}$.
We immediately have three
rules as (E) If $p_{e}\leq p_{r}$ and $p_{e}\leq p_{s_{\downarrow}}$, then $T_{\mu_{A}}(x)=1$;
(S) If $p_{s_{\downarrow}}\leq p_{e}$ and $p_{s_{\downarrow}}\leq p_{r}$, then $T_{\mu_{A}}(x)=0.5$;(R) If $p_{r}\leq p_{e}$ and $p_{r}\leq p_{s_{\downarrow}}$, then $T_{\mu_{A}}(x)=0$.

\begin{table}[htbp]\renewcommand{\arraystretch}{1.5}
\caption{The complementary matrix for situation 1.}
 \tabcolsep0.15in
\begin{tabular}{c c c c}
\hline  p & $\widetilde{R}_{e}(x)$ &$\widetilde{R}_{r}(x)$& $\widetilde{R}_{s_{\downarrow}}(x)$ \\\hline
$\widetilde{R}_{e}(x)$ & $p_{ee}=p(\widetilde{R}_{e}(x)\geq \widetilde{R}_{e}(x))$& $p_{er}=p(\widetilde{R}_{e}(x)\geq \widetilde{R}_{r}(x)) $ &$p_{es_{\downarrow}}=p(\widetilde{R}_{e}(x)\geq \widetilde{R}_{s_{\downarrow}}(x)) $ \\
$\widetilde{R}_{r}(x)$ & $p_{re}=p(\widetilde{R}_{r}(x)\geq \widetilde{R}_{e}(x))$& $p_{rr}=p(\widetilde{R}_{r}(x)\geq \widetilde{R}_{r}(x)) $  & $p_{rs_{\downarrow}}=p(\widetilde{R}_{r}(x)\geq \widetilde{R}_{s_{\downarrow}}(x)) $ \\
$\widetilde{R}_{s_{\downarrow}}(x)$  & $p_{s_{\downarrow}e}=p(\widetilde{R}_{s_{\downarrow}}(x)\geq \widetilde{R}_{e}(x))$& $p_{s_{\downarrow}r}=p(\widetilde{R}_{s_{\downarrow}}(x)\geq \widetilde{R}_{r}(x)) $   &$p_{s_{\downarrow}s_{\downarrow}}=p(\widetilde{R}_{s_{\downarrow}}(x)\geq \widetilde{R}_{s_{\downarrow}}(x)) $ \\
\hline
\end{tabular}
\end{table}

In consideration of Definition 6.1, $p_{er}=p(\widetilde{R}_{e}(x)\geq \widetilde{R}_{r}(x))$ has three kinds of possible results:
(I): $p_{er}=0$,
(II): $0<p_{er}<1,$
 and (III): $p_{er}=1$.
Furthermore, we have the similar results for $p_{es_{\downarrow}}$ and $p_{rs_{\downarrow}}$.

(1)
For $p_{er},$ if $p_{er}=0$, we have
\begin{eqnarray*}
\frac{m_{\theta}(\mu_{A}(x))\lambda^{+}_{r}-(1-m_{\theta}(\mu_{A}(x)))\lambda^{-}_{e}}{(1-m_{\theta}(\mu_{A}(x)))\lambda^{+}_{e}
-(1-m_{\theta}(\mu_{A}(x)))\lambda^{-}_{e}+m_{\theta}(\mu_{A}(x))\lambda^{+}_{r}
-m_{\theta}(\mu_{A}(x))\lambda^{-}_{r}}\geq1.
\end{eqnarray*}
In other words, we have
\begin{eqnarray*}
m_{\theta}(\mu_{A}(x))\lambda^{-}_{r}\geq (1-m_{\theta}(\mu_{A}(x)))\lambda^{+}_{e}\Leftrightarrow m_{\theta}(\mu_{A}(x))\geq\frac{\lambda^{+}_{e}}{\lambda^{-}_{r}+\lambda^{+}_{e}}.
\end{eqnarray*}

If $1>p_{er}>0$, we have
\begin{eqnarray*}
1>\frac{m_{\theta}(\mu_{A}(x))\lambda^{+}_{r}-(1-m_{\theta}(\mu_{A}(x)))\lambda^{-}_{e}}{(1-m_{\theta}(\mu_{A}(x)))\lambda^{+}_{e}
-(1-m_{\theta}(\mu_{A}(x)))\lambda^{-}_{e}+m_{\theta}(\mu_{A}(x))\lambda^{+}_{r}
-m_{\theta}(\mu_{A}(x))\lambda^{-}_{r}}>0.
\end{eqnarray*}
In other words, we have
\begin{eqnarray*}
(1-m_{\theta}(\mu_{A}(x)))\lambda^{+}_{e}&>&m_{\theta}(\mu_{A}(x))\lambda^{-}_{r}\Leftrightarrow m_{\theta}(\mu_{A}(x))<\frac{\lambda^{+}_{e}}{\lambda^{-}_{r}+\lambda^{+}_{e}};\\
m_{\theta}(\mu_{A}(x))\lambda^{+}_{r}&>&(1-m_{\theta}(\mu_{A}(x)))\lambda^{-}_{e}\Leftrightarrow m_{\theta}(\mu_{A}(x))<\frac{\lambda^{-}_{e}}{\lambda^{+}_{r}+\lambda^{-}_{e}}.
\end{eqnarray*}

If $p_{er}=1$, we have
\begin{eqnarray*}
\frac{m_{\theta}(\mu_{A}(x))\lambda^{+}_{r}-(1-m_{\theta}(\mu_{A}(x)))\lambda^{-}_{e}}{(1-m_{\theta}(\mu_{A}(x)))\lambda^{+}_{e}
-(1-m_{\theta}(\mu_{A}(x)))\lambda^{-}_{e}+m_{\theta}(\mu_{A}(x))\lambda^{+}_{r}
-m_{\theta}(\mu_{A}(x))\lambda^{-}_{r}}\leq0.
\end{eqnarray*}
In other words, we have
\begin{eqnarray*}
(1-m_{\theta}(\mu_{A}(x)))\lambda^{-}_{e}\geq m_{\theta}(\mu_{A}(x))\lambda^{+}_{r}
\Leftrightarrow m_{\theta}(\mu_{A}(x))\geq\frac{\lambda^{-}_{e}}{\lambda^{+}_{r}+\lambda^{-}_{e}}.
\end{eqnarray*}

(2)
For $p_{es_{\downarrow}},$ if $p_{es_{\downarrow}}=0$, we have
\begin{eqnarray*}
\frac{(m_{\theta}(\mu_{A}(x))-0.5)\lambda^{+}_{s_{\downarrow}}-(1-m_{\theta}(\mu_{A}(x)))\lambda^{-}_{e}}{(1-m_{\theta}(\mu_{A}(x)))\lambda^{+}_{e}
-(1-m_{\theta}(\mu_{A}(x)))\lambda^{-}_{e}+(m_{\theta}(\mu_{A}(x))-0.5)\lambda^{+}_{s_{\downarrow}}-(m_{\theta}(\mu_{A}(x))-0.5)\lambda^{-}_{s_{\downarrow}}}\geq1.
\end{eqnarray*}
In other words, we have
\begin{eqnarray*}
(m_{\theta}(\mu_{A}(x))-0.5)\lambda^{+}_{s_{\downarrow}}\geq (1-m_{\theta}(\mu_{A}(x)))\lambda^{+}_{e}
\Leftrightarrow m_{\theta}(\mu_{A}(x))\geq\frac{0.5\lambda^{-}_{s_{\downarrow}}+\lambda^{+}_{e}}{\lambda^{-}_{s_{\downarrow}}+\lambda^{+}_{e}}.
\end{eqnarray*}

If $1>p_{es_{\downarrow}}>0$, we have
\begin{eqnarray*}
1>\frac{(m_{\theta}(\mu_{A}(x))-0.5)\lambda^{+}_{s_{\downarrow}}-(1-m_{\theta}(\mu_{A}(x)))\lambda^{-}_{e}}{(1-m_{\theta}(\mu_{A}(x)))\lambda^{+}_{e}
-(1-m_{\theta}(\mu_{A}(x)))\lambda^{-}_{e}+(m_{\theta}(\mu_{A}(x))-0.5)\lambda^{+}_{s_{\downarrow}}-(m_{\theta}(\mu_{A}(x))-0.5)\lambda^{-}_{s_{\downarrow}}}>0.
\end{eqnarray*}
In other words, we have
\begin{eqnarray*}
(m_{\theta}(\mu_{A}(x))-0.5)\lambda^{+}_{s_{\downarrow}}&>&(1-m_{\theta}(\mu_{A}(x)))\lambda^{-}_{e}
\Leftrightarrow m_{\theta}(\mu_{A}(x))>\frac{0.5\lambda^{+}_{s_{\downarrow}}+\lambda^{-}_{e}}{\lambda^{-}_{e}+\lambda^{+}_{s_{\downarrow}}};
\\(1-m_{\theta}(\mu_{A}(x)))\lambda^{+}_{e}&>&(m_{\theta}(\mu_{A}(x))-0.5)\lambda^{+}_{s_{\downarrow}}
\Leftrightarrow m_{\theta}(\mu_{A}(x))<\frac{0.5\lambda^{-}_{s_{\downarrow}}+\lambda^{+}_{e}}{\lambda^{-}_{s_{\downarrow}}+\lambda^{+}_{e}}.
\end{eqnarray*}

If $p_{es_{\downarrow}}=1$, we have
\begin{eqnarray*}
\frac{(m_{\theta}(\mu_{A}(x))-0.5)\lambda^{+}_{s_{\downarrow}}-(1-m_{\theta}(\mu_{A}(x)))\lambda^{-}_{e}}{(1-m_{\theta}(\mu_{A}(x)))\lambda^{+}_{e}
-(1-m_{\theta}(\mu_{A}(x)))\lambda^{-}_{e}+(m_{\theta}(\mu_{A}(x))-0.5)\lambda^{+}_{s_{\downarrow}}-(m_{\theta}(\mu_{A}(x))-0.5)\lambda^{-}_{s_{\downarrow}}}\leq 0.
\end{eqnarray*}

In other words, we have
\begin{eqnarray*}
(m_{\theta}(\mu_{A}(x))-0.5)\lambda^{+}_{s_{\downarrow}}\leq(1-m_{\theta}(\mu_{A}(x)))\lambda^{-}_{e}
\Leftrightarrow m_{\theta}(\mu_{A}(x))\leq\frac{0.5\lambda^{+}_{s_{\downarrow}}+\lambda^{-}_{e}}{\lambda^{+}_{s_{\downarrow}}-\lambda^{+}_{r}}.
\end{eqnarray*}

(3)
For $p_{rs_{\downarrow}},$ if $p_{rs_{\downarrow}}=0$, we have
\begin{eqnarray*}
\frac{(m_{\theta}(\mu_{A}(x))-0.5)\lambda^{+}_{s_{\downarrow}}-m_{\theta}(\mu_{A}(x))\lambda^{-}_{r}}{m_{\theta}(\mu_{A}(x))\lambda^{+}_{r}
-m_{\theta}(\mu_{A}(x))\lambda^{-}_{r}+(m_{\theta}(\mu_{A}(x))-0.5)\lambda^{+}_{s_{\downarrow}}-(m_{\theta}(\mu_{A}(x))-0.5)\lambda^{-}_{s_{\downarrow}}}\geq1.
\end{eqnarray*}

In other words, we have
\begin{eqnarray*}
m_{\theta}(\mu_{A}(x))\lambda^{+}_{r}\leq(m_{\theta}(\mu_{A}(x))-0.5)\lambda^{-}_{s_{\downarrow}}
\Leftrightarrow m_{\theta}(\mu_{A}(x))\leq\frac{0.5\lambda^{-}_{s_{\downarrow}}}{\lambda^{-}_{s_{\downarrow}}-\lambda^{+}_{r}}.
\end{eqnarray*}

If $0<p_{rs_{\downarrow}}<1$, we have
\begin{eqnarray*}
1>\frac{(m_{\theta}(\mu_{A}(x))-0.5)\lambda^{+}_{s_{\downarrow}}-m_{\theta}(\mu_{A}(x))\lambda^{-}_{r}}{m_{\theta}(\mu_{A}(x))\lambda^{+}_{r}
-m_{\theta}(\mu_{A}(x))\lambda^{-}_{r}+(m_{\theta}(\mu_{A}(x))-0.5)\lambda^{+}_{s_{\downarrow}}-(m_{\theta}(\mu_{A}(x))-0.5)\lambda^{-}_{s_{\downarrow}}}>0.
\end{eqnarray*}

In other words, we have
\begin{eqnarray*}
(m_{\theta}(\mu_{A}(x))-0.5)\lambda^{+}_{s_{\downarrow}}&>&m_{\theta}(\mu_{A}(x))\lambda^{-}_{r}\Leftrightarrow m_{\theta}(\mu_{A}(x))<\frac{0.5\lambda^{+}_{s_{\downarrow}}}{\lambda^{+}_{s_{\downarrow}}-\lambda^{-}_{r}};\\
m_{\theta}(\mu_{A}(x))\lambda^{+}_{r}&>&(m_{\theta}(\mu_{A}(x))-0.5)\lambda^{-}_{s_{\downarrow}}
\Leftrightarrow m_{\theta}(\mu_{A}(x))>\frac{0.5\lambda^{-}_{s_{\downarrow}}}{\lambda^{+}_{s_{\downarrow}}-\lambda^{+}_{r}}.\end{eqnarray*}

If $p_{rs_{\downarrow}}=1$, we have
\begin{eqnarray*}
\frac{(m_{\theta}(\mu_{A}(x))-0.5)\lambda^{+}_{s_{\downarrow}}-m_{\theta}(\mu_{A}(x))\lambda^{-}_{r}}{m_{\theta}(\mu_{A}(x))\lambda^{+}_{e}
-m_{\theta}(\mu_{A}(x))\lambda^{-}_{e}+(m_{\theta}(\mu_{A}(x))-0.5)\lambda^{+}_{s_{\downarrow}}-(m_{\theta}(\mu_{A}(x))-0.5)\lambda^{-}_{s_{\downarrow}}}\leq0.
\end{eqnarray*}

In other words, we have
\begin{eqnarray*}
m_{\theta}(\mu_{A}(x))\lambda^{-}_{r}\geq(m_{\theta}(\mu_{A}(x))-0.5)\lambda^{+}_{s_{\downarrow}}
\Leftrightarrow m_{\theta}(\mu_{A}(x))\geq\frac{0.5\lambda^{+}_{s_{\downarrow}}}{\lambda^{+}_{s_{\downarrow}}-\lambda^{-}_{r}};
\end{eqnarray*}

\begin{table}[htbp]
\renewcommand{\arraystretch}{1.5}
\caption{Types of operations for situation 1} \tabcolsep0.5in
\begin{tabular}{ccccccccccccc}
\hline
$$ &$p_{er}$ & $p_{es_{\downarrow}}$ &
$p_{rs_{\downarrow}}$&$T_{\mu_{A}}(x)$ \\\hline
$1$& I & I & I &  1 \\
$2$& I & I & II &  1 \\
$3$& I & I & III &  1 \\
$4$& I & II & I &  1 \\
$5$& I & II & II &  1\text{ or } 0.5  \\
$6$& I & II & III & 1\text{ or } 0.5   \\
$7$& I & III & I &  1 \\
$8$& I & III & II &  0.5  \\
$9$& I & III & III &  0.5  \\
$10$& II & I & I &    1\text{ or } 0 \\
$11$& II & I & II &   1\text{ or } 0 \text{ or } 0.5\\
$12$& II & I & III &   1 \\
$13$& II & II & I &    1\text{ or } 0\\
$14$& II & II & II &    1\text{ or } 0.5\text{ or } 0\\
$15$& II & II & III &  1\text{ or } 0.5  \\
$16$& II & III & I &   0 \\
$17$& II & III & II &   0.5\text{ or } 0 \\
$18$& II & III & III &  0.5  \\
$19$& III & I & I &    0\\
$20$& III & I & II &    0\\
$21$& III & I & III &  1 \\
$22$& III & II & I &   0 \\
$23$& III & II & II &   0.5\text{ or } 0 \\
$24$& III & II & III &  0.5  \\
$25$& III & III & I &   0 \\
$26$& III & III & II &  0.5\text{ or } 0  \\
$27$& III & III & III &  0.5  \\
\hline
\end{tabular}
\end{table}

\begin{table}[htbp]
\renewcommand{\arraystretch}{1.5}
\caption{Special types of operations for situation 1.}
\tabcolsep0.5in
\begin{tabular}{ccc}
\hline
$\text{ Type }$ &$\text{ Condition }$ &$T_{\mu_{A}}(x)$ \\\hline
$5$& $2p_{es_{\downarrow}}+p_{rs_{\downarrow}}\leq 2$ & 1 \\
   & $2p_{es_{\downarrow}}+p_{rs_{\downarrow}}> 2$ & 0.5 \\
$6$& $p_{es_{\downarrow}}\leq \frac{1}{2}$ & 0 \\
   & $p_{es_{\downarrow}}> \frac{1}{2}$ & 0.5 \\
$10$& $p_{er}\leq \frac{1}{2}$ & 1 \\
    & $p_{er}> \frac{1}{2}$ & 0 \\
$11$& $2p_{es_{\downarrow}}-p_{rs_{\downarrow}}\leq 1$ & 1 \\
    & $2p_{es_{\downarrow}}-p_{rs_{\downarrow}}> 1$ & 0 \\
$13$& $2p_{er}+p_{rs_{\downarrow}}\leq 1\wedge p_{er}+2p_{es_{\downarrow}}\leq 2$ & 1 \\
    & $2p_{er}+p_{rs_{\downarrow}}\leq1\wedge -p_{er}+2p_{es_{\downarrow}}\leq 1$ & 0 \\
$14$& $ 2p_{er}+p_{es_{\downarrow}}-p_{rs_{\downarrow}}\leq1\wedge p_{er}+2p_{es_{\downarrow}}+p_{rs_{\downarrow}}\leq 2$ & 1 \\
    & $ 1\leq2p_{er}+p_{es_{\downarrow}}-p_{rs_{\downarrow}}\leq1\wedge p_{es_{\downarrow}}-p_{er}+2p_{rs_{\downarrow}}\leq 1$ & 0 \\
    & $ 2\leq p_{er}+2p_{es_{\downarrow}}+p_{rs_{\downarrow}}\wedge 1\leq p_{es_{\downarrow}}-p_{er}+2p_{rs_{\downarrow}}\leq 1$ & 0 \\
$15$&  $2p_{er}+p_{es_{\downarrow}}\leq 2\wedge p_{er}+2p_{rs_{\downarrow}}\leq 1$ & 1 \\
    &  $1\leq2p_{er}+p_{es_{\downarrow}}\wedge p_{er}-p_{es_{\downarrow}}\leq 1$ & 0.5 \\
$17$&  $2p_{rs_{\downarrow}}- p_{er}\leq 0$ & 0 \\
    &  $2p_{rs_{\downarrow}}- p_{er}> 0$ & 0.5 \\
$23$& $2p_{rs_{\downarrow}}+ p_{es_{\downarrow}}\leq 2$ & 0 \\
    & $2p_{rs_{\downarrow}}+ p_{er}> 2$ & 0.5 \\
$26$& $p_{rs_{\downarrow}}\leq 0.5$ & 0 \\
    & $p_{rs_{\downarrow}}>0.5$ & 0.5 \\
\hline
\end{tabular}
\end{table}

\begin{example} (Continuation of Example 3.3)
Let $\widetilde{\lambda}_{e}=[1,2],\widetilde{\lambda}_{r}=[5,6]$ and $
\widetilde{\lambda}_{s_{\downarrow}}=[3,4]$ when $m_{\theta}(\mu_{A}(x))\geq 0.5,$ we have that
\begin{eqnarray*}
\widetilde{R}_{e}(x_{2})&=&\widetilde{\lambda}_{e}E_{e}(m_{\theta}(\mu_{A}(x_{2})))=[0.3\lambda^{-}_{e},0.3\lambda^{+}_{e}]=[0.3,0.6];\\
\widetilde{R}_{r}(x_{2})&=&\widetilde{\lambda}_{r}E_{r}(m_{\theta}(\mu_{A}(x_{2})))=[0.7\lambda^{-}_{r},0.7\lambda^{+}_{r}]=[3.5,4.2];
\\
\widetilde{R}_{s_{\downarrow}}(x_{2})&=&\widetilde{\lambda}_{s_{\downarrow}}E_{s_{\downarrow}}(m_{\theta}(\mu_{A}(x_{2})))=[0.2\lambda^{-}_{s_{\downarrow}},0.2\lambda^{+}_{s_{\downarrow}}]=[0.6,0.8].
\end{eqnarray*}

In light of the complementary matrix $P_{ers_{\downarrow}}$,
we have
\begin{eqnarray*}
P_{ers_{\downarrow}}=\left[
\begin{array}{ccc}
p_{ee}  & p_{er}  & p_{es_{\downarrow}} \\
p_{re}  & p_{rr}  & p_{rs_{\downarrow}}\\
p_{s_{\downarrow}e}  & p_{s_{\downarrow}r}  & p_{s_{\downarrow}s_{\downarrow}}\\
\end{array}
\right]=\left[
\begin{array}{ccc}
0.5  & 0  & 0 \\
1  & 0.5  & 1\\
1  & 0  & 0.5\\
\end{array}
\right].
\end{eqnarray*}

Therefore, we have that elevating the membership grade of $x_{2}$ to 1 is the best choice.
\end{example}

\subsection{Situation 2: $m_{\theta}(\mu_{A}(x))< 0.5$}

 For $m_{\theta}(\mu_{A}(x))\leq 0.5$, we have
\begin{eqnarray*}
P_{ers_{\uparrow}}=\left[
\begin{array}{ccc}
p_{ee}  & p_{er}  & p_{es_{\uparrow}} \\
p_{re}  & p_{rr}  & p_{rs_{\uparrow}}\\
p_{s_{\uparrow}e}  & p_{s_{\uparrow}r}  & p_{s_{\uparrow}s_{\uparrow}}\\
\end{array}
\right].
\end{eqnarray*}

According to the properties of the degree of possibilities, we have
$p_{ee}=p_{rr}=p_{s_{\uparrow}s_{\uparrow}}=0.5,p_{er}+p_{re}=1,
p_{es_{\uparrow}}+p_{s_{\uparrow}e}=1$ and $ p_{rs_{\uparrow}}+p_{s_{\uparrow}r}=1.$ Then we simplify the matrix as
\begin{eqnarray*}
P_{ers_{\uparrow}}=\left[
\begin{array}{ccc}
0.5  & p_{er}  & p_{es_{\uparrow}} \\
1-p_{er}  & 0.5  & p_{rs_{\uparrow}}\\
1-p_{es_{\uparrow}}  & 1-p_{rs_{\uparrow}}  & 0.5\\
\end{array}
\right].
\end{eqnarray*}

In light of the complementary matrix $P_{ers_{\uparrow}}$, all elements in each
line of the matrix are summarized as
\begin{eqnarray*}
p_{e}=0.5+p_{er}+ p_{es_{\uparrow}};
p_{r}=0.5-p_{er}+p_{rs_{\uparrow}};
p_{s_{\uparrow}}=2.5-p_{es_{\uparrow}}-p_{rs_{\uparrow}},
\end{eqnarray*}
where $p_{e}$ is the total degree of preference of $\widetilde{R}_{e}(x);$
$p_{r}$ is the total degree of preference of $\widetilde{R}_{r}(x);$
$p_{s_{\uparrow}}$ is the total degree of preference of $\widetilde{R}_{s_{\uparrow}}(x)$. The values of $p_{e},p_{r}$ and $p_{s_{\uparrow}}$ depend on $p_{er},p_{es_{\uparrow}}$ and $p_{rs_{\uparrow}}$.
We immediately have three
rules as (E) If $p_{e}\leq p_{r}$ and $p_{e}\leq p_{s_{\uparrow}}$, then $T_{\mu_{A}}(x)=1$;
(S) If $p_{s_{\uparrow}}\leq p_{e}$ and $p_{s_{\uparrow}}\leq p_{r}$, then $T_{\mu_{A}}(x)=0.5$;(R) If $p_{r}\leq p_{e}$ and $p_{r}\leq p_{s_{v}}$, then $T_{\mu_{A}}(x)=0$.

\begin{table}[htbp]\renewcommand{\arraystretch}{1.5}
\caption{The complementary matrix for situation 2.}
 \tabcolsep0.15in
\begin{tabular}{c c c c}
\hline  p & $\widetilde{R}_{e}(x)$ &$\widetilde{R}_{r}(x)$& $\widetilde{R}_{s_{\uparrow}}(x)$ \\\hline
$\widetilde{R}_{e}(x)$ & $p_{ee}=p(\widetilde{R}_{e}(x)\geq \widetilde{R}_{e}(x))$& $p_{er}=p(\widetilde{R}_{e}(x)\geq \widetilde{R}_{r}(x)) $ &$p_{es_{\uparrow}}=p(\widetilde{R}_{e}(x)\geq \widetilde{R}_{s_{\uparrow}}(x)) $ \\
$\widetilde{R}_{r}(x)$ & $p_{re}=p(\widetilde{R}_{r}(x)\geq \widetilde{R}_{e}(x))$& $p_{rr}=p(\widetilde{R}_{r}(x)\geq \widetilde{R}_{r}(x)) $  & $p_{rs_{\uparrow}}=p(\widetilde{R}_{r}(x)\geq \widetilde{R}_{s_{\uparrow}}(x)) $ \\
$\widetilde{R}_{s_{\uparrow}}(x)$  & $p_{s_{\uparrow}e}=p(\widetilde{R}_{s_{\uparrow}}(x)\geq \widetilde{R}_{e}(x))$& $p_{s_{\uparrow}r}=p(\widetilde{R}_{s_{\uparrow}}(x)\geq \widetilde{R}_{r}(x)) $   &$p_{s_{\uparrow}s_{\uparrow}}=p(\widetilde{R}_{s_{\uparrow}}(x)\geq \widetilde{R}_{s_{\uparrow}}(x)) $ \\
\hline
\end{tabular}
\end{table}

In consideration of Definition 6.1, $p_{er}=p(\widetilde{R}_{e}(x)\geq \widetilde{R}_{r}(x))$ have three kinds of possible results:
(I): $p_{er}=0$,
(II): $0<p_{er}<1;$
(III): $p_{er}=1$.
Furthermore, we have the similar results for $p_{es_{\downarrow}}$ and $p_{rs_{\downarrow}}$.

(1)
For $p_{er},$ if $p_{er}=0$, we have
\begin{eqnarray*}
\frac{m_{\theta}(\mu_{A}(x))\lambda^{+}_{r}-(1-m_{\theta}(\mu_{A}(x)))\lambda^{-}_{e}}{(1-m_{\theta}(\mu_{A}(x)))\lambda^{+}_{e}
-(1-m_{\theta}(\mu_{A}(x)))\lambda^{-}_{e}+m_{\theta}(\mu_{A}(x))\lambda^{+}_{r}
-m_{\theta}(\mu_{A}(x))\lambda^{-}_{r}}\geq1.
\end{eqnarray*}
In other words, we have
\begin{eqnarray*}
m_{\theta}(\mu_{A}(x))\lambda^{-}_{r}
\geq (1-m_{\theta}(\mu_{A}(x)))\lambda^{+}_{e}\Leftrightarrow m_{\theta}(\mu_{A}(x))\geq\frac{\lambda^{+}_{e}}{\lambda^{-}_{r}+\lambda^{+}_{e}}.
\end{eqnarray*}

If $0<p_{er}<1$, we have
\begin{eqnarray*}
1>\frac{m_{\theta}(\mu_{A}(x))\lambda^{+}_{r}-(1-m_{\theta}(\mu_{A}(x)))\lambda^{-}_{e}}{(1-m_{\theta}(\mu_{A}(x)))\lambda^{+}_{e}
-(1-m_{\theta}(\mu_{A}(x)))\lambda^{-}_{e}+m_{\theta}(\mu_{A}(x))\lambda^{+}_{r}
-m_{\theta}(\mu_{A}(x))\lambda^{-}_{r}}>0.
\end{eqnarray*}
In other words, we have
\begin{eqnarray*}
m_{\theta}(\mu_{A}(x))\lambda^{-}_{r} &>& (1-m_{\theta}(\mu_{A}(x)))\lambda^{+}_{e}\Leftrightarrow m_{\theta}(\mu_{A}(x))>\frac{\lambda^{+}_{e}}{\lambda^{+}_{e}+\lambda^{-}_{r}};
\\
m_{\theta}(\mu_{A}(x))\lambda^{+}_{r}&>&(1-m_{\theta}(\mu_{A}(x)))\lambda^{-}_{e}\Leftrightarrow m_{\theta}(\mu_{A}(x))>\frac{\lambda^{-}_{e}}{\lambda^{+}_{r}+\lambda^{-}_{e}}.
\end{eqnarray*}

If $p_{er}=1$, we have
\begin{eqnarray*}
\frac{m_{\theta}(\mu_{A}(x))\lambda^{+}_{r}-(1-m_{\theta}(\mu_{A}(x)))\lambda^{-}_{e}}{(1-m_{\theta}(\mu_{A}(x)))\lambda^{+}_{e}
-(1-m_{\theta}(\mu_{A}(x)))\lambda^{-}_{e}+m_{\theta}(\mu_{A}(x))\lambda^{+}_{r}
-m_{\theta}(\mu_{A}(x))\lambda^{-}_{r}}\leq0.
\end{eqnarray*}
In other words, we have
\begin{eqnarray*}
m_{\theta}(\mu_{A}(x))\lambda^{+}_{r}\leq(1-m_{\theta}(\mu_{A}(x)))\lambda^{-}_{e}
\Leftrightarrow m_{\theta}(\mu_{A}(x))\leq\frac{\lambda^{-}_{e}}{\lambda^{+}_{r}+\lambda^{-}_{e}}.
\end{eqnarray*}

(2)
For $p_{es_{\uparrow}},$ if $p_{es_{\uparrow}}=0$, we have
\begin{eqnarray*}
\frac{(0.5-m_{\theta}(\mu_{A}(x)))\lambda^{+}_{s_{\uparrow}}-(1-m_{\theta}(\mu_{A}(x)))\lambda^{-}_{e}}{(1-m_{\theta}(\mu_{A}(x)))\lambda^{+}_{e}
-(1-m_{\theta}(\mu_{A}(x)))\lambda^{-}_{e}+(0.5-m_{\theta}(\mu_{A}(x)))\lambda^{+}_{s_{\uparrow}}-(0.5-m_{\theta}(\mu_{A}(x)))\lambda^{-}_{s_{\uparrow}}}\geq1.
\end{eqnarray*}
In other words, we have
\begin{eqnarray*}
(0.5-m_{\theta}(\mu_{A}(x)))\lambda^{+}_{s_{\uparrow}}\geq (1-m_{\theta}(\mu_{A}(x)))\lambda^{+}_{e}
\Leftrightarrow m_{\theta}(\mu_{A}(x))\geq\frac{0.5\lambda^{-}_{s_{\uparrow}}-\lambda^{+}_{e}}{\lambda^{-}_{s_{\uparrow}}-\lambda^{+}_{e}}.
\end{eqnarray*}

If $1>p_{es_{\uparrow}}>0$, we have
\begin{eqnarray*}
1>\frac{(0.5-m_{\theta}(\mu_{A}(x)))\lambda^{+}_{s_{\uparrow}}-(1-m_{\theta}(\mu_{A}(x)))\lambda^{-}_{e}}{(1-m_{\theta}(\mu_{A}(x)))\lambda^{+}_{e}
-(1-m_{\theta}(\mu_{A}(x)))\lambda^{-}_{e}+(0.5-m_{\theta}(\mu_{A}(x)))\lambda^{+}_{s_{\uparrow}}-(0.5-m_{\theta}(\mu_{A}(x)))\lambda^{-}_{s_{\uparrow}}}>0.
\end{eqnarray*}
In other words, we have
\begin{eqnarray*}
(0.5-m_{\theta}(\mu_{A}(x)))\lambda^{+}_{s_{\uparrow}}&>&(1-m_{\theta}(\mu_{A}(x)))\lambda^{-}_{e}\Leftrightarrow m_{\theta}(\mu_{A}(x))>\frac{\lambda^{-}_{e}-0.5\lambda^{+}_{s_{\uparrow}}}{\lambda^{-}_{e}-\lambda^{+}_{s_{\uparrow}}};
\\
(1-m_{\theta}(\mu_{A}(x)))\lambda^{+}_{e}&>&(0.5-m_{\theta}(\mu_{A}(x)))\lambda^{+}_{s_{\uparrow}}\Leftrightarrow m_{\theta}(\mu_{A}(x))>\frac{0.5\lambda^{-}_{s_{\uparrow}}-\lambda^{-}_{e}}{\lambda^{-}_{s_{\uparrow}}-\lambda^{+}_{e}}.
\end{eqnarray*}

If $p_{es_{\uparrow}}=1$, we have
\begin{eqnarray*}
\frac{(0.5-m_{\theta}(\mu_{A}(x)))\lambda^{+}_{s_{\uparrow}}-(1-m_{\theta}(\mu_{A}(x)))\lambda^{-}_{e}}{(1-m_{\theta}(\mu_{A}(x)))\lambda^{+}_{e}
-(1-m_{\theta}(\mu_{A}(x)))\lambda^{-}_{e}+(0.5-m_{\theta}(\mu_{A}(x)))\lambda^{+}_{s_{\uparrow}}-(0.5-m_{\theta}(\mu_{A}(x)))\lambda^{-}_{s_{\uparrow}}}\leq 0.
\end{eqnarray*}

In other words, we have
\begin{eqnarray*}
(1-m_{\theta}(\mu_{A}(x)))\lambda^{+}_{e}\geq(0.5-m_{\theta}(\mu_{A}(x)))\lambda^{+}_{s_{\uparrow}}
\Leftrightarrow m_{\theta}(\mu_{A}(x))\leq\frac{0.5\lambda^{+}_{s_{\uparrow}}-\lambda^{-}_{e}}{\lambda^{+}_{s_{\uparrow}}-\lambda^{-}_{e}}.
\end{eqnarray*}

(3)
For $p_{rs_{\uparrow}},$ if $p_{rs_{\uparrow}}=0$, we have
\begin{eqnarray*}
\frac{(0.5-m_{\theta}(\mu_{A}(x)))\lambda^{+}_{s_{\uparrow}}-m_{\theta}(\mu_{A}(x))\lambda^{-}_{r}}{m_{\theta}(\mu_{A}(x))\lambda^{+}_{r}
-m_{\theta}(\mu_{A}(x))\lambda^{-}_{r}+(0.5-m_{\theta}(\mu_{A}(x)))\lambda^{+}_{s_{\uparrow}}-(0.5-m_{\theta}(\mu_{A}(x)))\lambda^{-}_{s_{\uparrow}}}\geq1.
\end{eqnarray*}

In other words, we have
\begin{eqnarray*}
m_{\theta}(\mu_{A}(x))\lambda^{+}_{r}\leq(0.5-m_{\theta}(\mu_{A}(x)))\lambda^{-}_{s_{\uparrow}}
\Leftrightarrow m_{\theta}(\mu_{A}(x))\leq\frac{0.5\lambda^{-}_{s_{\uparrow}}}{\lambda^{-}_{s_{\uparrow}}+\lambda^{+}_{r}}.
\end{eqnarray*}

If $1>p_{rs_{\downarrow}}>0$, we have
\begin{eqnarray*}
1>\frac{(0.5-m_{\theta}(\mu_{A}(x)))\lambda^{+}_{s_{\uparrow}}-m_{\theta}(\mu_{A}(x))\lambda^{-}_{r}}{m_{\theta}(\mu_{A}(x))\lambda^{+}_{r}
-m_{\theta}(\mu_{A}(x))\lambda^{-}_{r}+(0.5-m_{\theta}(\mu_{A}(x)))\lambda^{+}_{s_{\uparrow}}-(0.5-m_{\theta}(\mu_{A}(x)))\lambda^{-}_{s_{\uparrow}}}>0.
\end{eqnarray*}

In other words, we have
\begin{eqnarray*}
(0.5-m_{\theta}(\mu_{A}(x)))\lambda^{+}_{s_{\uparrow}}&>&m_{\theta}(\mu_{A}(x))\lambda^{-}_{r}\Leftrightarrow m_{\theta}(\mu_{A}(x))<\frac{0.5\lambda^{+}_{s_{\uparrow}}}{\lambda^{+}_{s_{\uparrow}}+\lambda^{-}_{r}};
\\
(0.5-m_{\theta}(\mu_{A}(x)))\lambda^{-}_{s_{\uparrow}}&>&m_{\theta}(\mu_{A}(x))\lambda^{+}_{r}\Leftrightarrow m_{\theta}(\mu_{A}(x))>\frac{0.5\lambda^{-}_{s_{\uparrow}}}{\lambda^{-}_{s_{\uparrow}}+\lambda^{+}_{r}}.
\end{eqnarray*}

If $p_{rs_{\downarrow}}=1$, we have
\begin{eqnarray*}
\frac{(0.5-m_{\theta}(\mu_{A}(x)))\lambda^{+}_{s_{\uparrow}}-m_{\theta}(\mu_{A}(x))\lambda^{-}_{r}}{(1-m_{\theta}(\mu_{A}(x)))\lambda^{+}_{r}
-(1-m_{\theta}(\mu_{A}(x)))\lambda^{-}_{r}+(0.5-m_{\theta}(\mu_{A}(x)))\lambda^{+}_{s_{\uparrow}}-(0.5-m_{\theta}(\mu_{A}(x)))\lambda^{-}_{s_{\uparrow}}}\leq0.
\end{eqnarray*}

In other words, we have
\begin{eqnarray*}
m_{\theta}(\mu_{A}(x))\lambda^{-}_{r}\geq(0.5-m_{\theta}(\mu_{A}(x)))\lambda^{+}_{s_{\uparrow}}\Leftrightarrow m_{\theta}(\mu_{A}(x))\geq\frac{0.5\lambda^{+}_{s_{\uparrow}}}{\lambda^{-}_{r}+\lambda^{+}_{s_{\uparrow}}}.
\end{eqnarray*}

\begin{table}[htbp]
\renewcommand{\arraystretch}{1.5}
\caption{Types of operations for situation 2.} \tabcolsep0.5in
\begin{tabular}{ccccccccccccc}
\hline
$$ &$p_{er}$ & $p_{es_{\uparrow}}$ &
$p_{rs_{\uparrow}}$&$T_{\mu_{A}}(x)$ \\\hline
$1$& I & I & I &  1 \\
$2$& I & I & II &  1 \\
$3$& I & I & III &  1 \\
$4$& I & II & I &  1 \\
$5$& I & II & II &  1\text{ or } 0.5  \\
$6$& I & II & III & 1\text{ or } 0.5   \\
$7$& I & III & I &  1 \\
$8$& I & III & II &  0.5  \\
$9$& I & III & III &  0.5  \\
$10$& II & I & I &    1\text{ or } 0 \\
$11$& II & I & II &   1\text{ or } 0 \text{ or } 0.5\\
$12$& II & I & III &   1 \\
$13$& II & II & I &    1\text{ or } 0\\
$14$& II & II & II &    1\text{ or } 0.5\text{ or } 0\\
$15$& II & II & III &  1\text{ or } 0.5  \\
$16$& II & III & I &   0 \\
$17$& II & III & II &   0.5\text{ or } 0 \\
$18$& II & III & III &  0.5  \\
$19$& III & I & I &    0\\
$20$& III & I & II &    0\\
$21$& III & I & III &  1 \\
$22$& III & II & I &   0 \\
$23$& III & II & II &   0.5\text{ or } 0 \\
$24$& III & II & III &  0.5  \\
$25$& III & III & I &   0 \\
$26$& III & III & II &  0.5\text{ or } 0  \\
$27$& III & III & III &  0.5  \\
\hline
\end{tabular}
\end{table}

\begin{table}[htbp]
\renewcommand{\arraystretch}{1.5}
\caption{Special types of operations for situation 2.}
\tabcolsep0.5in
\begin{tabular}{ccc}
\hline
$\text{ Type }$ &$\text{ Condition }$ &$T_{\mu_{A}}(x)$ \\\hline
$5$& $2p_{es_{\uparrow}}+p_{rs_{\uparrow}}\leq 2$ & 1 \\
   & $2p_{es_{\uparrow}}+p_{rs_{\uparrow}}> 2$ & 0.5 \\
$6$& $p_{es_{\uparrow}}\leq \frac{1}{2}$ & 0 \\
   & $p_{es_{\uparrow}}> \frac{1}{2}$ & 0.5 \\
$10$& $p_{er}\leq \frac{1}{2}$ & 1 \\
    & $p_{er}> \frac{1}{2}$ & 0 \\
$11$& $2p_{es_{\uparrow}}-p_{rs_{\uparrow}}\leq 1$ & 1 \\
    & $2p_{es_{\uparrow}}-p_{rs_{\uparrow}}> 1$ & 0 \\
$13$& $2p_{er}+p_{rs_{\uparrow}}\leq 1\wedge p_{er}+2p_{es_{\uparrow}}\leq 2$ & 1 \\
    & $2p_{er}+p_{rs_{\uparrow}}\leq1\wedge -p_{er}+2p_{es_{\uparrow}}\leq 1$ & 0 \\
$14$& $ 2p_{er}+p_{es_{\uparrow}}-p_{rs_{\uparrow}}\leq1\wedge p_{er}+2p_{es_{\uparrow}}+p_{rs_{\uparrow}}\leq 2$ & 1 \\
    & $ 1\leq2p_{er}+p_{es_{\uparrow}}-p_{rs_{\uparrow}}\leq1\wedge p_{es_{\uparrow}}-p_{er}+2p_{rs_{\downarrow}}\leq 1$ & 0 \\
    & $ 2\leq p_{er}+2p_{es_{\downarrow}}+p_{rs_{\downarrow}}\wedge 1\leq p_{es_{\uparrow}}-p_{er}+2p_{rs_{\uparrow}}\leq 1$ & 0 \\
$15$&  $2p_{er}+p_{es_{\uparrow}}\leq 2\wedge p_{er}+2p_{rs_{\uparrow}}\leq 1$ & 1 \\
    &  $1\leq2p_{er}+p_{es_{\uparrow}}\wedge p_{er}-p_{es_{\uparrow}}\leq 1$ & 0.5 \\
$17$&  $2p_{rs_{\uparrow}}- p_{er}\leq 0$ & 0 \\
    &  $2p_{rs_{\uparrow}}- p_{er}> 0$ & 0.5 \\
$23$& $2p_{rs_{\uparrow}}+ p_{es_{\uparrow}}\leq 2$ & 0 \\
    & $2p_{rs_{\uparrow}}+ p_{er}> 2$ & 0.5 \\
$26$& $p_{rs_{\uparrow}}\leq 0.5$ & 0 \\
    & $p_{rs_{\uparrow}}>0.5$ & 0.5 \\
\hline
\end{tabular}
\end{table}

\begin{example} (Continuation of Example 3.3)
Let $\widetilde{\lambda}_{e}=[5,6],\widetilde{\lambda}_{r}=[1,2]$ and $
\widetilde{\lambda}_{s_{\uparrow}}=[3,4]$ when $m_{\theta}(\mu_{A}(x))< 0.5,$ we have
\begin{eqnarray*}
\widetilde{R}_{e}(x_{3})&=&\widetilde{\lambda}_{e}E_{e}(m_{\theta}(\mu_{A}(x_{3})))=[0.6\lambda^{-}_{e},0.6\lambda^{+}_{e}]=[3,3.6];\\
\widetilde{R}_{r}(x_{3})&=&\widetilde{\lambda}_{r}E_{r}(m_{\theta}(\mu_{A}(x_{3})))=[0.4\lambda^{-}_{r},0.4\lambda^{+}_{r}]=[0.4,0.8];
\\
\widetilde{R}_{s_{\uparrow}}(x_{3})&=&\widetilde{\lambda}_{s_{\uparrow}}E_{s_{\uparrow}}(m_{\theta}(\mu_{A}(x_{3})))=[0.1\lambda^{-}_{s_{\uparrow}},0.1\lambda^{+}_{s_{\uparrow}}]=[0.3,0.4].
\end{eqnarray*}

In light of the complementary matrix $P_{ers_{\uparrow}}$,
we have
\begin{eqnarray*}
P_{ers_{\uparrow}}=\left[
\begin{array}{ccc}
p_{ee}  & p_{er}  & p_{es_{\uparrow}} \\
p_{re}  & p_{rr}  & p_{rs_{\uparrow}}\\
p_{s_{\uparrow}e}  & p_{s_{\uparrow}r}  & p_{s_{\uparrow}s_{\uparrow}}\\
\end{array}
\right]=\left[
\begin{array}{ccc}
0.5  & 1  & 1 \\
0  & 0.5  & 1\\
0  & 0  & 0.5\\
\end{array}
\right].
\end{eqnarray*}

Therefore, we have that elevating the membership of $x_{3}$ to 0.5 is the best choice.
\end{example}

\section{Four semantics issues of this model}

In this section, we investigate four semantics issues of
decision-theoretic three-way approximations of interval-valued fuzzy sets.

(1) Interpretations of interval-valued loss functions

In decision-theoretic three-way approximations of interval-valued fuzzy
sets, the pair of thresholds depends on the choice of interval-valued loss
functions which are fundamental notions of the decision-theoretic
model. In other words, given an interval-valued loss function, the pair of
thresholds can be computed accordingly. On the other hand, if the
pair of thresholds is interpreted in terms of an interval-valued loss
function, then the user can provide a better estimation of the
thresholds in time. Therefore, the decision-theoretic model gives an
interpretation of the pair of thresholds, and it is important to
discuss approximations of interval-valued fuzzy sets by using interval-valued loss
functions.

(2) Relationships to shadowed sets of fuzzy sets

In Sections 3 and 4, we see that three regions of decision-theoretic
rough set-based three-way approximation $T_{\mu_{(A)}}$ and shadowed
set $S_{\mu_{(A)}}$ are both defined through a pair of thresholds
$\alpha$ and $\beta$. For shadowed sets, the objective function is
given with respect to the membership functions, and different
membership functions will produce different shadowed sets. In
contrast, the objective function of the decision-theoretic framework
is given with respect to interval-valued loss functions, which is
independent of any particular fuzzy membership functions.

(3) Relationships to decision-theoretic rough sets

In Section 4, we adopt the main ideas from decision-theoretic rough
set in developing decision-theoretic rough set-based three-way
approximations of interval-valued fuzzy sets. A rough membership function
can be viewed as a fuzzy membership function. There are some
differences between three-way approximations of interval-valued fuzzy sets
and decision-theoretic rough sets. For decision-theoretic rough
sets, we deal with two-state three-way decision problems. A rough
membership function denotes the probability that an object is in the
set. On the other hand, three-way approximations of interval-valued fuzzy
sets are a many-state decision problem.

(4) Relationships to decision-theoretic three-way approximations of
fuzzy sets

In \cite{Deng1}, Deng et al. discussed three-way approximations of
fuzzy sets by using loss functions. In practice, there are a lot of
interval-valued loss functions. Compare with Deng's model, we discuss
three-way approximations of interval-valued fuzzy sets by using interval-valued
loss functions.

\section{Conclusions}

Many researchers have investigated approximations of interval-valued fuzzy sets. In
this paper, firstly, we have presented shadowed sets for
interpreting and understanding interval-valued fuzzy sets. Secondly, we have
constructed decision-theoretic rough set-based three-way
approximations of interval-valued fuzzy sets. Thirdly, we have computed the
pair of thresholds for decision-theoretic rough set-based three-way
approximations of interval-valued fuzzy sets by using interval-valued loss
functions. Fourthly, we have constructed approximations of interval-valued fuzzy sets by using interval-valued loss functions from another view. Finally, we have employed several examples to illustrate
that how to make a decision for interval-valued fuzzy sets by using
interval-valued loss functions.

There are still many interesting topics deserving further
investigations on fuzzy sets. For example, there are many types of
fuzzy sets and loss functions, and it is of interest to
investigate loss functions-based three-way approximations of
interval-valued fuzzy sets. In the future, we will further investigate
interval-valued fuzzy sets and discuss its application in knowledge
discovery.

\section*{ Acknowledgments}

We would like to thank the anonymous reviewers very much for their
professional comments and valuable suggestions. This work is
supported by the National Natural Science Foundation of China (NO.
11371130, 11401052, 11401195), the Scientific Research Fund of Hunan
Provincial Education Department (No.14C0049).


\begin{thebibliography}{00}

\bibitem{Banerjee1}
M. Banerjee, S.K. Pal, Roughness of a fuzzy set, Information
Sciences 93 (1996) 235-246.\vskip.10in

\bibitem{Cattaneo1}
G. Cattaneo, D. Ciucci, Shadowed sets and related algebraic
structures, Fundamenta Informaticae 55 (2003) 255-284.\vskip.10in

\bibitem{Cattaneo2}
G. Cattaneo, D. Ciucci, An algebraic approach to shadowed sets,
Electronic Notes in Theoretical Computer Science 82 (2003) 64-75.
\vskip.10in

\bibitem{Cattaneo3}
G. Cattaneo, D. Ciucci, Theoretical aspects of shadowed sets, W.
Pedrycz, A. Skowron, V. Kreinovich (Eds.), Handbook of Granular
Computing, John Wiley and Sons, New York (2008), pp.
603-628.\vskip.10in

\bibitem{Chakrabarty1}
K. Chakrabarty, R. Biswas, S. Nanda, Nearest ordinary set of a fuzzy
set: a rough theoretic construction, Bulletin of the Polish Academy
of Sciences: Technical Sciences  46 (1998) 105-114.\vskip.10in

\bibitem{Chanas1}
S. Chanas, On the interval approximation of a fuzzy number, Fuzzy
Sets and Systems 122 (2001) 353-356.\vskip.10in

\bibitem{Deng1}
X.F. Deng, Y.Y. Yao, Decision-theoretic three-way approximations of
fuzzy sets, Information Sciences  279 (2014) 702-715. \vskip.10in

\bibitem{Deng2}
X.F. Deng, Y.Y. Yao, Mean-value-based decision-theoretic shadowed
sets, W. Pedrycz, M.Z. Reformat (Eds.), Proceedings of 2013 Joint
IFSA World Congress and NAFIPS Annual Meeting (IFSA/NAFIPS), IEEE
Press, New York (2013) 1382-1387.\vskip.10in

\bibitem{Dubois1}
D. Dubois, H. Prade, Fuzzy Sets and Fuzzy Rough Sets: Theory and
Applications, Academic Press, New York (1980).\vskip.10in

\bibitem{Grzegorzewski1}
P. Grzegorzewski, Nearest interval approximation of a fuzzy number,
Fuzzy Sets and Systems 130 (2002) 321-330.\vskip.10in

\bibitem{Grzegorzewski2}
P. Grzegorzewski, Fuzzy number approximation via shadowed sets,
Information Sciences 225 (2013) 35-46.\vskip.10in

\bibitem{Klir1}
G.J. Klir, Y. Bo, Fuzzy Sets and Fuzzy Logic: Theory and
Applications, Prentice Hall, New Jersey (1995).\vskip.10in

\bibitem{Li1}
H.X. Li, X.Z. Zhou, Risk decision making based on decision-theoretic
rough set: a three-way view decision model, International Journal of
Computational Intelligence Systems 4 (2011) 1-11.\vskip.10in

\bibitem{Liang1}
D.C. Liang, D. Liu, W. Pedrycz, P. Hu, Triangular fuzzy
decision-theoretic rough sets, International Journal of Approximate
Reasoning 54 (2013) 1087-1106.\vskip.10in

\bibitem{Liang2}
D.C. Liang, D. Liu, Systematic studies on three-way decisions with
interval-valued decision-theoretic rough sets, Information Sciences
276 (2014) 186-203.\vskip.10in

\bibitem{Liu1}
D. Liu, T.R. Li, D.C. Liang, Interval-valued decision-theoretic
rough sets, Computer Science 39(7) (2012) 178-181. \vskip.10in

\bibitem{Liu2}
D. Liu, Y.Y. Yao, T.R. Li, Three-way investment decisions with
decision-theoretic rough sets, International Journal of
Computational Intelligence Systems 4 (2011) 66-74.\vskip.10in

\bibitem{Liu3}
D. Liu, T.R. Li, D.C. Liang, Three-Way Decisions in Dynamic
Decision-Theoretic Rough Sets, Lecture Notes in Computer Science
8171 (2013) 291-301.\vskip.10in

\bibitem{Mitra1}
S. Mitra, P.P. Kundu, Satellite image segmentation with shadowed
C-means, Information Sciences 181 (2011) 3601-3613.\vskip.10in

\bibitem{Moore1}
R. Moore, W. Lodwick, Interval analysis and fuzzy set theory, Fuzzy Sets and Systems
135 (2003) 5¨C9.\vskip.10in

\bibitem{Nasibov1}
E.N. Nasibov, S. Peker, On the nearest parametric approximation of a
fuzzy number, Information Sciences 159 (2008) 1365-1375.\vskip.10in


\bibitem{Pedrycz1}
W. Pedrycz, Shadowed sets: representing and processing fuzzy sets,
IEEE Transactions on Systems Man Cybernetics-Systems 28 (1998)
103-109.\vskip.10in

\bibitem{Pedrycz2}
W. Pedrycz, Shadowed sets: bridging fuzzy and rough sets, S.K. Pal,
A. Skowron (Eds.), Rough Fuzzy Hybridization: A New Trend in
Decision-Making, Springer, Singapore (1999) 179-199.\vskip.10in

\bibitem{Pedrycz3}
W. Pedrycz, Fuzzy clustering with a knowledge-based guidance,
Pattern Recognition Letters 25 (2004) 469-480.\vskip.10in

\bibitem{Pedrycz4}
W. Pedrycz, Interpretation of clusters in the framework of shadowed
sets, Pattern Recognition Letters 26 (2005) 2439-2449.\vskip.10in

\bibitem{Pedrycz5}
W. Pedrycz, From fuzzy sets to shadowed sets: Interpretation and
computing, International Journal of Intelligent Systems 24 (2009)
 48-61.\vskip.10in

\bibitem{Pedrycz6}
A. Pedrycz, F. Dong, K. Hirota, Finite cut-based approximation of
fuzzy sets and its evolutionary optimization, Fuzzy Sets and Systems
160 (2009) 3550-3564.\vskip.10in

\bibitem{Qian1}
Y.H. Qian, J.Y. Liang, C.Y. Dang, Interval ordered information systems, Computers and Mathematics with Applications
 56 (2008) 1994-2009.\vskip.10in

\bibitem{Sengupta1}
A. Sengupta, T.K. Pal, On comparing interval numbers, European Journal of Operational Research
 127 (2000) 28-43.\vskip.10in

\bibitem{Slezak1}
D. Slezak, W. Ziarko, The investigation of the Bayesian rough set
model, International Journal of Approximate Reasoning 40 (2005)
81-91.\vskip.10in


\bibitem{Turksen1}
I.B. Turksen, Interval valued fuzzy sets based on normal forms,
Fuzzy Sets and Systems, 20(2) (1986) 191-210.\vskip.10in

\bibitem{Wang1}
L. Wang, J. Wang, Feature weighting fuzzy clustering integrating
rough sets and shadowed sets, International Journal of Pattern
Recognition and Artificial Intelligence 26 (2012)
http://dx.doi.org/10.1142/S0218001412500103.\vskip.10in

\bibitem{Yao1}
Y.Y. Yao, Probabilistic rough set approximations, International
Journal of Approximate Reasoning 49 (2008) 255-271.\vskip.10in

\bibitem{Yao2}
Y.Y. Yao, Three-way decisions with probabilistic rough sets,
Information Sciences 180 (2010) 341-353.\vskip.10in


\bibitem{Yao3}
Y.Y. Yao, Two semantic issues in a probabilistic rough set model,
Fundamenta Informaticae 108 (2011) 249-265. \vskip.10in

\bibitem{Yao4}
Y.Y. Yao, Probabilistic approaches to rough sets, Expert Systems 20
(2003) 287-297.\vskip.10in

\bibitem{Yao5}
Y.Y. Yao, The superiority of three-way decision in probabilistic
rough set models, Information Sciences 181 (6) (2011)
1080-1096.\vskip.10in

\bibitem{Ziarko1}
W. Ziarko, Probabilistic approach to rough set, International
Journal of Approximate Reasoning 49 (2008) 272-284.\vskip.10in

\bibitem{Zadeh1}
L.A. Zadeh, Fuzzy sets, Information and Control 8 (1965)
338-353.\vskip.10in

\bibitem{Zadeh2}
L.A. Zadeh, Is there a need for fuzzy logic? Information Sciences
178 (2008) 2751-2779.\vskip.10in

\bibitem{Zhou1}
J. Zhou, W. Pedryczb, D.Q. Miao, Shadowed sets in the
characterization of rough-fuzzy clustering, Pattern Recognition 44
(2011) 1738-1749.\vskip.10in


\end{thebibliography}
\end{document}